%% file: main.tex
\documentclass{sbc2023}%
\usepackage[caption=false,farskip=0pt]{subfig}

\usepackage{newtxtext}
\usepackage{newtxmath}
\usepackage{color}
\usepackage{aas_macros}
\usepackage[bottom]{footmisc}
\usepackage{supertabular}
\usepackage{afterpage}
\usepackage{url}
\usepackage{multicol}
\usepackage{multirow}
\usepackage[acronym]{glossaries}
\usepackage[table,dvipsnames]{xcolor}
\usepackage{xspace}
\usepackage[capitalise,noabbrev]{cleveref}
\usepackage{hyperref}
\usepackage{cleveref}
\setcitestyle{square}
\usepackage{arydshln}
\usepackage{tabularx}
\usepackage{verbatimbox}
\usepackage{graphicx}
\definecolor{orcidlogo}{rgb}{0.37,0.48,0.13}
\definecolor{unilogo}{rgb}{0.16, 0.26, 0.58}
\definecolor{maillogo}{rgb}{0.58, 0.16, 0.26}
\definecolor{darkblue}{rgb}{0.0,0.0,0.0}
\hypersetup{colorlinks,breaklinks,
            linkcolor=darkblue,urlcolor=darkblue,
            anchorcolor=darkblue,citecolor=darkblue}
\usepackage{tabularray}
\usepackage{booktabs}
\usepackage{balance}
\usepackage{multirow}
\usepackage{makecell}

\usepackage{ifthen}
\usepackage[normalem]{ulem} %
\usepackage{stackengine}

\newacronymstyle{long-short-br}
{%
  \GlsUseAcrEntryDispStyle{long-short}%
}%
{%
  \GlsUseAcrStyleDefs{long-short}%
}
\setacronymstyle{long-short-br}

\jtitle{This is an author-prepared version of a paper accepted for publication in the \textit{Journal of the Brazilian Computer Society}.\\
The final version will be available at the SBC-OpenLib Library.}

\title{Toward Unified Fine-Grained Vehicle Classification and Automatic License Plate Recognition}

\author{

\affil{\textbf{Gabriel E. Lima}\textsuperscript{*}~~[~\textbf{Federal University of Paraná~}|{~\textbf{\textit{gelima@inf.ufpr.br}}}~]}

\affil{\textbf{Valfride Nascimento}~~[~\textbf{Federal University of Paraná}~|\href{mailto:vwnascimento@inf.ufpr.br}{~\textbf{\textit{vwnascimento@inf.ufpr.br}}}~]}

\affil{\textbf{Eduardo Santos}~~[~\textbf{Paraná Military Police, Federal University of Paraná}~|\href{mailto:ed.santos@pm.pr.gov.br}{~\textbf{\textit{ed.santos@pm.pr.gov.br}}}~]}

\affil{\textbf{Eduil Nascimento Jr.}~~[~\textbf{Paraná Military Police}~|\href{mailto:eduiljunior@pm.pr.gov.br}{~\textbf{\textit{eduiljunior@pm.pr.gov.br}}}~]}

\affil{\textbf{Rayson Laroca}~~[~\textbf{Pontifical Catholic University of Paraná, Federal University of Paraná}~|\href{mailto:rayson@ppgia.pucpr.br}{~\textbf{\textit{rayson@ppgia.pucpr.br}}}~]}

\affil{\textbf{David Menotti}~~[~\textbf{Federal University of Paraná~}|\href{mailto:menotti@inf.ufpr.br}{~\textbf{\textit{menotti@inf.ufpr.br}}}~]}

}

\usepackage{ifthen}
\usepackage[dvipsnames,table]{xcolor}

\newcommand\major[1]{#1} %

\newboolean{showcomments}
\setboolean{showcomments}{true} %
\newboolean{showcommentsMA}
\setboolean{showcommentsMA}{true} %
\usepackage{footmisc}
\newcommand\red[1]{{\textcolor{red}{#1}}}

\newcommand*{\MyComment}[4][]{
  \ifthenelse{\boolean{showcomments}}{%
    \textcolor{#2}{[\textbf{\ifthenelse{\equal{#1}{}}{#3}{#3(#1)}}: #4]}%
  }{}%
}

\newcommand{\NEW}[1]{\major{#1}}
\newcommand{\REWRITTEN}[1]{\major{#1}}

\newcounter{fncounter}
\begin{document}

\input{0-acronyms}

\begin{frontmatter}
\maketitle

\begin{mail}
Department of Informatics, Federal University of Paraná, R. Evaristo F. Ferreira da Costa 391, Jardim das Américas, Curitiba, PR, 81530-090, Brazil. 
\end{mail}

\input{0-abstract}

\end{frontmatter}

\input{1-intro}
\input{2-related}
\input{3-dataset}
\input{4-analysis}
\input{5-fine-grained}
\input{6-alpr}
\input{7-conclusion}

\input{0-acknowledgments}

\section*{Declarations}

\begin{contributions}
Gabriel E. Lima is the primary contributor and writer of the manuscript. 
Valfride Nascimento assisted with experiment validation and participated in the manuscript review. Eduardo Santos contributed to data selection and was involved in the manuscript review process. 
Eduil Nascimento Jr.\ provided the image resources and contributed to the manuscript review. 
Rayson Laroca co-advised the project, supporting its conceptualization and methodology, and contributed to the review and editing of the manuscript.
David Menotti supervised the project, provided resources, and contributed to the review and editing of the manuscript.
All authors read and approved the final~manuscript.
\end{contributions}

\begin{interests}
The authors declare that they have no competing interests.
\end{interests}

\begin{materials}
The resources generated and analyzed during this study are available at \textbf{\url{https://github.com/Lima001/UFPR-VeSV-Dataset}}.
\end{materials}

\bibliographystyle{apalike-sol}
\bibliography{bibtex}
\balance

\end{document}

%% file: 0-acronyms.tex
\newacronym{alpr}{ALPR}{Automatic License Plate Recognition}
\newacronym{cnn}{CNN}{Convolutional Neural Network}
\newacronym{fgvc}{FGVC}{Fine-Grained Vehicle Classification}
\newacronym{its}{ITS}{Intelligent Transportation Systems}
\newacronym{lpd}{LPD}{License Plate Detection}
\newacronym{lpr}{LPR}{License Plate Recognition}
\newacronym{senatran}{SENATRAN}{Brazilian National Traffic Secretariat}
\newacronym{sibgrapi}{SIBGRAPI}{Conference on Graphics, Patterns, and Images}
\newacronym{srr}{SRR}{Softmax Response Rejection}
\newacronym{spp}{SPP}{Spatial Pyramid Pooling}
\newacronym{ocr}{OCR}{Optical Character Recognition}
\newacronym{vcr}{VCR}{Vehicle Color Recognition}
\newacronym{vmmr}{VMMR}{Vehicle Make and Model Recognition}
\newacronym{vtr}{VTR}{Vehicle Type Recognition}
\newacronym{vit}{ViT}{Vision Transformer}

\newcommand{\rodosolalpr}{RodoSol-ALPR\xspace}
\newcommand{\ufpralpr}{UFPR-ALPR\xspace}
\newcommand{\dataset}{UFPR-VeSV\xspace}

\newacronym{lp}{LP}{License Plate}

%% file: 0-abstract.tex
\begin{abstract} 
\textbf{Abstract.} 
Extracting vehicle information from surveillance images is essential for intelligent transportation systems, enabling applications such as traffic monitoring and criminal investigations. 
While \gls*{alpr} is widely used, \gls*{fgvc} offers a complementary approach by identifying vehicles based on attributes such as color, make, model, and type. 
Although there have been advances in this field, existing studies often assume well-controlled conditions, explore limited attributes, and overlook \gls*{fgvc} integration with \gls*{alpr}.  
To address these gaps, we introduce \dataset, a dataset comprising $24{,}945$ images of \major{$16{,}297$} unique vehicles with annotations for $13$ colors, $26$ makes, $136$ models, and $14$ types. 
Collected from the Military Police of Paraná (Brazil) surveillance system, the dataset captures diverse real-world conditions, including partial occlusions, nighttime infrared imaging, and varying lighting. 
All \gls*{fgvc} annotations were validated using license plate information, with text and corner annotations also being provided.  
\NEW{A qualitative and quantitative comparison with established datasets confirmed the challenging nature of our dataset. 
A benchmark using five deep learning models further validated this, revealing specific challenges such as handling multicolored vehicles, infrared images, and distinguishing between vehicle models that share a common platform.} 
Additionally, we apply two optical character recognition models to license plate recognition and explore the joint use of \gls*{fgvc} and \gls*{alpr}. 
The results highlight the potential of integrating these complementary tasks for real-world applications. 
The \dataset dataset is publicly available at:~\textbf{\url{https://github.com/Lima001/UFPR-VeSV-Dataset}}.
\end{abstract}

\begin{keywords}
Intelligent Transportation Systems, Fine-Grained Vehicle Classification, Automatic License Plate Recognition, Surveillance.
\end{keywords}

%% file: 1-intro.tex
\glsresetall

\section{Introduction}
\label{sec:intro}

Extracting vehicle information from surveillance images is a crucial aspect of \gls*{its}, enabling applications such as traffic monitoring and criminal  investigations~\citep{yang2015compcars,he2024vehicle,laroca2025advancing}. 
Traditional vehicle identification methods primarily rely on \gls*{alpr}. 
However, \gls*{alpr} techniques are susceptible to partial occlusions, viewpoint variations, and poor image quality, all of which can degrade recognition performance~\citep{fan2022improving,nascimento2024enhancing,wojcik2025lplc}.

\gls*{fgvc} presents a complementary solution by categorizing vehicles based on attributes such as color, make, model, type, and year.
Unlike \glspl*{lp}, these attributes are often more resilient to occlusions and viewpoint changes.
Additionally, although distinguishing visually similar vehicles is an inherently complex task, recent research has achieved remarkable results~\citep{wang2020multipath,lu2023efficient}.

However, a review of the literature reveals that \gls*{fgvc} research assumes at least one of the following controlled conditions: fixed viewpoints, sufficient lighting, or high-quality images.
These assumptions fail to fully capture the challenges of real-world surveillance scenarios.
Moreover, existing research frequently addresses make and model recognition separately from color and type recognition~\citep{amirkhani2023deepcar,hu2023vehicle}, hindering a comprehensive exploration of vehicle~attributes.

Another unexplored aspect in existing research is the integration of \gls*{fgvc} and \gls*{alpr}, which could offer significant benefits~\citep{oliveira2021vehicle}. 
By cross-referencing \gls*{fgvc} attributes with \gls*{lp} records, \gls*{alpr} system errors can be identified, reducing false positives and enhancing information retrieval.
Furthermore, \gls*{fgvc} can help refine the search space for \gls*{alpr}, particularly in forensic scenarios where \glspl*{lp} are low-quality or occluded \citep{nascimento2025toward}.

To bridge these gaps, we present the publicly available UFPR Vehicle Surveillance~(\dataset) dataset, consisting of $24{,}945$ images of \major{$16{,}297$} unique vehicles.
It includes annotations across 13 color classes, 26 makes, 136 models, and 14 vehicle types.
Sourced from the Military Police of Paraná (Brazil) surveillance system, the dataset reflects diverse real-world conditions, including frontal and rear views, partial occlusions, varying lighting, and nighttime infrared imaging. 
\glspl*{lp} were manually annotated and used to retrieve vehicle information for validating the \gls*{fgvc} annotations.
The dataset also includes annotations for both \gls*{lp} characters and corner~coordinates.

To establish the value of the \dataset dataset, this paper makes the following contributions.
First, we conduct a comprehensive comparison with existing datasets to highlight the dataset’s novelty. 
Second, we empirically validate its challenging nature by showing that a simple transfer-learning approach, effective on related datasets, fails to achieve adequate results on ours. 
Third, we establish performance baselines by benchmarking five deep learning models for \gls*{fgvc} tasks and evaluating two optical character recognition models for \gls*{lpr}. 
Finally, as a step toward a unified solution, we explore the integration of \gls*{fgvc} and \gls*{alpr} and analyse its effectiveness for enhancing vehicle information retrieval.

The dataset's design, which prioritized high-quality \gls*{fgvc} annotations, resulted in the exclusion of truly unconstrained \gls*{alpr} scenarios~(see \cref{subsec:annotation_process}). 
This limitation likely explains the high \gls*{lpr} performance, which surpassed individual \gls*{fgvc} tasks. 
Despite this observation, the experiments confirm that integrating \gls*{alpr} and \gls*{fgvc} is a promising strategy for enhancing vehicle information retrieval. 
This joint analysis represents a significant step toward a unified approach, and the challenges identified herein provide a foundation for future research.

A preliminary version of this research was published at the 2024 \gls*{sibgrapi}~\citep{lima2024toward}. 
This work expands upon that study in several key aspects. 
First, we broaden the scope beyond vehicle color recognition to include the recognition of vehicle make, model, and type. 
We also introduce a larger and more diverse dataset that more accurately reflects real-world conditions. 
To highlight its contributions, we compare the proposed dataset with prominent existing ones in \gls*{alpr} and \gls*{fgvc}. 
Additionally, the methodology has been enhanced through the integration of an additional deep-learning model, along with improvements in training and evaluation procedures.
Lastly, we present \gls*{lpr} results on the new dataset and introduce a novel contribution by integrating our best-performing \gls*{lpr} and \gls*{fgvc} methods.

The remainder of this paper is organized as follows. 
\cref{sec:RelatedWork} reviews related work. 
\cref{sec:Dataset} introduces the \dataset dataset. 
\cref{sec:ComparativeAnalysis} provides a comparative analysis with existing datasets. 
\cref{sec:FGVC} presents the experimental evaluation of \gls*{fgvc} tasks. 
\cref{sec:ALPR} details the experimental evaluation of \gls*{lpr} and its subsequent integration with \gls*{fgvc}. 
Finally, \cref{sec:Conclusion} concludes the paper and discusses future research directions.

%% file: 2-related.tex
\section{Related Work}
\label{sec:RelatedWork}

This section provides an overview of recent studies and datasets in the fields of \gls*{fgvc}  and \gls*{alpr}. 
\cref{subsec:rw_vcr}  focuses on vehicle color recognition, while \cref{subsec:rw_vmmr} covers vehicle make and model recognition, and \cref{subsec:rw_vtr} is related to vehicle type recognition. 
\cref{subsec:rw_alpr} reviews \gls*{alpr}-related research. 
A differentiation of existing studies and this work is presented separately in~\cref{subsec:differentiators}.

\subsection{Vehicle Color Recognition}
\label{subsec:rw_vcr}

Vehicle color recognition plays a significant role in vehicle identification, as color information is visually distinctive, less affected by occlusions, and remains stable across viewpoints~\citep{chen2014vehicle}.
Early studies used small datasets captured in controlled environments, relying on handcrafted feature extraction methods combined with machine learning classifiers for color prediction~\citep{baek2007vehicle, son2007convolution, dule2010convenient}. 
These works laid the groundwork for more robust and advanced methodologies.

The first large-scale, publicly available dataset for this task was introduced by~\cite{chen2014vehicle}, comprising $15{,}601$ frontal-view images categorized into eight colors.
Their initial approach employed a region-of-interest selector and support vector machines.
Due to its diverse conditions -- including variations in lighting, haze, and overexposure -- the dataset became a popular benchmark for subsequent research using \gls*{cnn} models~\citep{fu2018mcff,zhang2018vehicle,hu2015vehicle}.

Following that, researchers have introduced new datasets with new scenarios to be explored.
\cite{wang2021transformer} proposed a dataset of $32{,}220$ rear-view images, classified into 11~colors and 75~subcategories, and explored a hybrid \gls*{cnn}-\gls*{vit} model for recognition. 
\cite{hu2023vehicle} introduced the Vehicle Color-24 dataset, consisting of $31{,}232$ frontal-view images distributed across 24~color classes, and developed a CNN with multi-scale feature fusion and a specialized loss function to address class imbalance, reaching promising~results.

More recently, in~\citep{lima2024toward}, we introduced UFPR-VCR, a dataset designed to capture real-world challenges such as partial occlusions and nighttime conditions. 
It consists of $10{,}039$ images taken under varying conditions, including both frontal and rear views. 
By evaluating four deep learning architectures, we achieved a peak accuracy of $66.2$\%, highlighting the complexity of color recognition in unconstrained environments and underscoring the importance of investigating such~scenarios.

\subsection{Vehicle Make and Model Recognition}
\label{subsec:rw_vmmr}

Vehicle make and model recognition is a challenging task in \gls*{fgvc} due to its high intra-class variability and low inter-class differences~\citep{wang2020multipath,oliveira2021vehicle}. 
Early research in this area began with the Stanford Cars-196 dataset~\citep{krause2013collecting,krause20133d}, which comprises $16{,}185$ web-sourced images from $196$ vehicle models. 
The initial benchmark, achieved using the BubbleBank algorithm~\citep{deng2013bubble}, was later surpassed by deep learning-based approaches~\citep{yu2022embedding,lu2023efficient}, which have since become the standard in the~field.

To better reflect real-world conditions, \cite{yang2015compcars} introduced CompCars, a dataset containing both web-sourced and surveillance images. 
Its surveillance subset~(CompCars-SV) includes $44{,}481$ frontal-view images of $281$ vehicle models annotated with multiple attributes. 
Similarly, \cite{sochor2016boxcars} developed the BoxCars dataset, consisting of $63{,}750$ surveillance images with multi-viewpoint data and 3D bounding box annotations, further challenging the recognition of vehicle make and model.

\cite{wang2020multipath} introduced MPF-Cars, a large-scale dataset with $335{,}011$ images from $2{,}019$ models and $180$ manufacturers. 
Their proposed three-branch \gls*{cnn} model leveraged full-vehicle, front, and logo views to improve identification performance. 
Likewise, \cite{kuhn2021brcars} presented BRCars, a dataset of $300{,}325$ images covering $427$ car models from online advertisements. 
Differently from related works, the dataset includes both exterior and interior view images, with recognition performed without distinguishing between the~two.

Recent research has focused on advancing recognition methods and scenarios. 
\cite{amirkhani2023deepcar} introduced the DeepCar~5.0 dataset, utilizing \glspl*{cnn} to analyze vehicle headlights, grilles, and bumpers for enhanced recognition. 
Additionally, \cite{wolf2024knowledge} explored an open-set recognition scenario and proposed a knowledge-distillation-based label smoothing approach, improving both closed-set and open-set recognition on the CompCars-SV dataset.

\subsection{Vehicle Type Recognition}
\label{subsec:rw_vtr}

Vehicle type recognition provides a coarse-grained classification level when compared to vehicle make and model recognition, distinguishing between categories such as cars, trucks, and motorcycles.
Early studies used handcrafted feature extraction methods applied to small and limited datasets~\citep{ferryman1995generic,jollyvehicle1996,lai2001vehicle,wu2001method,ma2005edge}.

\cite{dong2015vehicle} pioneered the use of deep learning in recognizing vehicle type and introduced BIT-Vehicle, a dataset of $9{,}850$ high-resolution frontal-view images from six types. 
\cite{hu2017location-aware} later proposed a multi-task \gls*{cnn} for joint vehicle localization and type recognition, and presented the SYSU-Vehicle dataset with $5{,}000$ web-sourced images and five classes. 
Later, \cite{shvai2020accurate} expanded the field by compiling a dataset of $73{,}638$ toll booth images and integrating \gls*{cnn} features with optical sensor data.

Recent studies have explored unconventional scenarios to improve recognition under challenging conditions. 
For example, \cite{basak2024vehicle} employed residual dense networks to generate super-resolved images, enhancing accuracy in low-resolution images. 
In a different approach, \cite{luo2024dense_tnt} developed a method for satellite imagery, combining DenseNet~\citep{huang2017densenet} and Transformer-in-Transformer~\citep{han2021tnt} layers to extract fine-grained spatial features.

\subsection{Automatic License Plate Recognition}
\label{subsec:rw_alpr}

\gls*{alpr} systems typically comprise two main components: \gls*{lpd} and \gls*{lpr}~\citep{laroca2023leveraging,laroca2025advancing}. 
\gls*{lpd} identifies the license plate region within an image, while \gls*{lpr} extracts and interprets the characters. 
One of the earliest widely adopted datasets within the area is AOLP~\citep{hsu2013aolp}, which includes $2{,}049$ images across three subsets, each tailored to different \gls*{alpr}~applications.

Further research aimed to improve evaluation scenarios proposing datasets such as \major{PKU~\citep{yuan2017robust}, SSIG-ALPR~\citep{goncalves2018realtime}, and UFPR-ALPR~\citep{laroca2018robust}}. 
Nevertheless, \cite{xu2018towards} identified limitations within those datasets in either scale (containing fewer than $10{,}000$ images) or diversity.
This led to the creation of CCPD, a large-scale dataset with $250{,}000$ images captured by roadside parking management personnel using handheld~cameras.

Further expanding dataset diversity, \cite{laroca2022cross} introduced RodoSol-ALPR, a dataset of $20{,}000$ toll booth images collected under varying conditions, including different times of day, weather scenarios, and camera distances. 
It was the first publicly available dataset within the field to include Mercosur license plates.
Additionally, it remains the largest dataset with annotated motorcycle images.

Recent research has focused on improving feature extraction to better handle generalization in real-world scenarios. 
\cite{rao2024license} and \cite{liu2024irregular} utilized spatial attention mechanisms to enhance both \gls*{lp} detection and recognition. 
Moreover,  \cite{liu2024improving} explored the use of synthetically generated data to increase dataset diversity, demonstrating its effectiveness in enhancing recognition~performance.

Lastly, it is worth pointing out the work from \cite{oliveira2021vehicle}, which introduced the Vehicle-Rear dataset. 
It contains $3{,}000$ rear-view images with extra annotations for vehicle color, make, and model. 
The authors proposed a dual-stream \gls*{cnn} network and integrated vehicle appearance features with \gls*{lpr} to perform vehicle identification. 
Despite including annotations suitable for both \gls*{alpr} and \gls*{fgvc} tasks, the dataset was not explored for the~latter.

\subsection{Key Differentiators and Contributions}
\label{subsec:differentiators}

This work's primary contribution is the introduction of the \dataset dataset, designed to capture vehicles under challenging surveillance conditions often absent in existing benchmarks. 
Beyond its capture conditions, the dataset introduces novel complexities for \gls*{fgvc} tasks. 
For color recognition, it includes scenarios rarely addressed in the literature, such as infrared images and multicolored vehicles. 
For type recognition, a finer class granularity is employed to distinguish between challenging categories (e.g., motorcycles versus scooters). 
For model recognition, the dataset includes vehicles that are visually similar due to shared manufacturing platforms, even across different vehicle types. 
The dataset also incorporates rear-view images of trucks with obstructed bodies and motorcycles. 

Finally, our research methodology sets this work apart. 
First, we not only evaluate \gls*{fgvc} tasks in isolation but also assess the methods for vehicle color, make, model, and type recognition jointly, revealing specific challenges in that combined context.
Second, we benchmark \gls*{lpr} methods on our proposed dataset. Finally, we conduct a specific analysis integrating the outputs of both \gls*{lpr} and \gls*{fgvc} models. 
This unified approach allows us to investigate recognition challenges under adverse conditions and showcase how these complementary systems can be combined.

%% file: 3-dataset.tex
\section{\dataset Dataset}
\label{sec:Dataset}

\dataset comprises $24{,}945$ images of \major{$16{,}297$} unique vehicles, specifically designed to support \gls*{fgvc} research in real-world surveillance scenarios.
It encompasses vehicles classified into $13$ colors, $26$ makes, $136$ models, and $14$ types.
The dataset exhibits a highly unbalanced distribution across these attributes (see \cref{fig:fgvc_class_distribution}), reflecting the characteristics of Brazilian traffic~\citep{gov2024frota,farias2023colorido,celestino2021marcas}. 
This poses a challenge for recognition models, which often struggle with underrepresented classes~\citep{huang2016learning,ochal2023fewshot}.

The images were sourced from the Military Police of Paraná’s surveillance system within a single municipality, captured by distinct cameras positioned on highways, urban streets, and rural roads. 
The cameras operate under diverse conditions, capturing vehicles from varying angles and distances, with environmental factors such as lighting, weather, and motion blur influencing image quality. 
Additionally, approximately $3$\% of the images were manually captured by police officers during monitoring operations, introducing further data~variety.

Although the dataset is designed to focus on a single vehicle per image, additional vehicles may appear due to camera positioning and perspective, as shown in \cref{fig:multiple_vehicles}. 
In some cases, multiple \glspl*{lp} are visible but partially occluded.
These scenarios are retained to reflect real-world conditions and the challenges they pose for \gls*{alpr} and \gls*{fgvc}-based~systems.

\setcounter{figure}{1}
\begin{figure}[!htb]
    \centering
        \includegraphics[height=17ex]{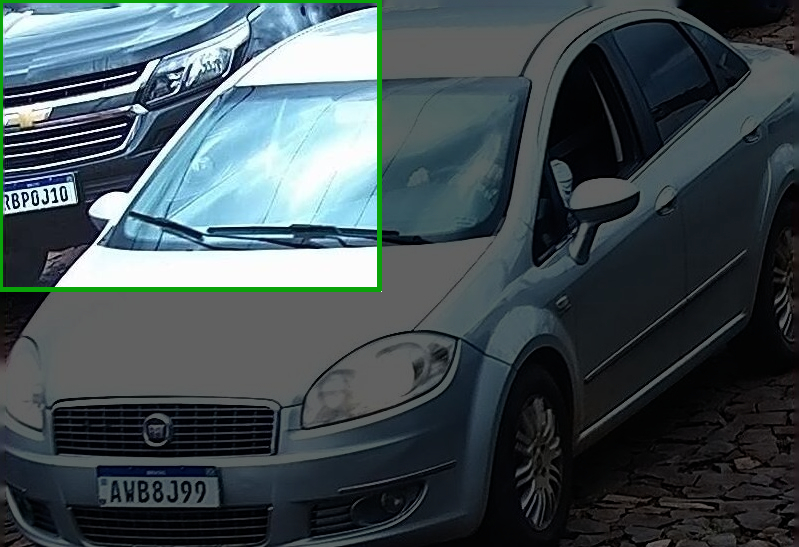}
        \includegraphics[height=17ex]{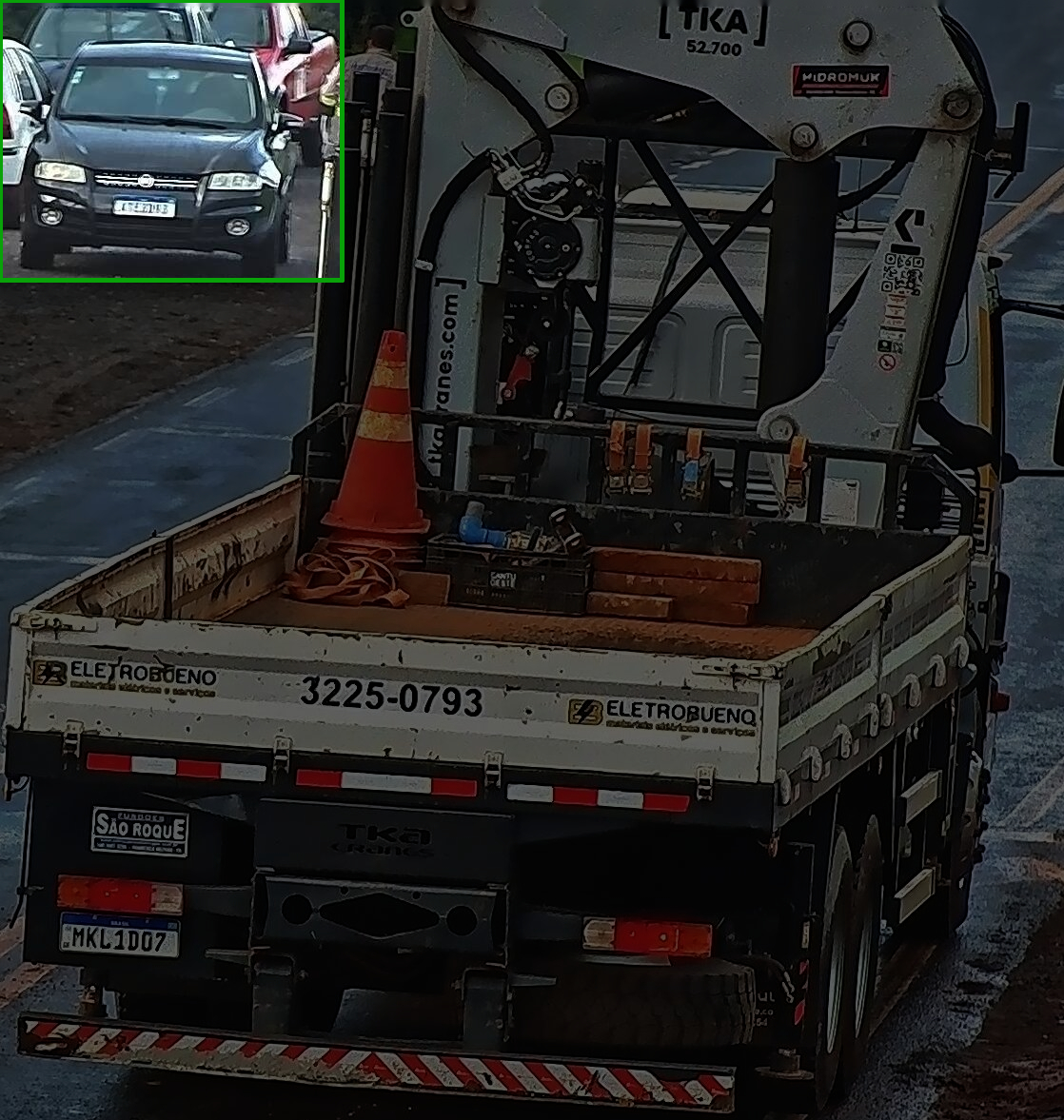}

    \vspace{0.5mm}
    
    \caption{Examples of images featuring multiple vehicles due to camera perspective. The background vehicle is highlighted with a green border, while the main vehicle is shadowed to enhance contrast.}
    \label{fig:multiple_vehicles}
\end{figure}

\begin{figure}[!htb]
    \centering
    \captionsetup[subfigure]{captionskip=1.5pt,font=scriptsize}
    \resizebox{0.995\linewidth}{!}{
        \subfloat[Adequate]{\includegraphics[height=12ex]{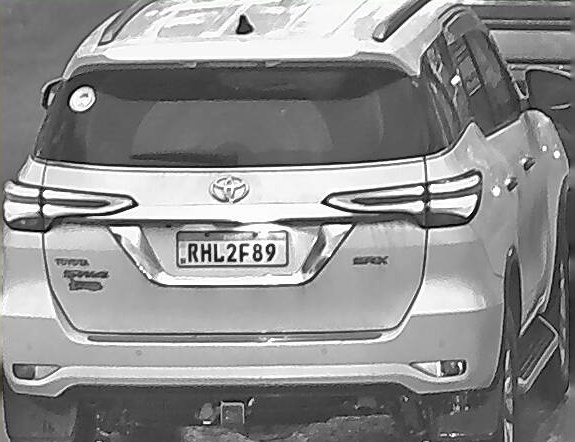}}
        \hspace{0.2mm}
        \subfloat[Reduced contrast]{\includegraphics[height=12ex]{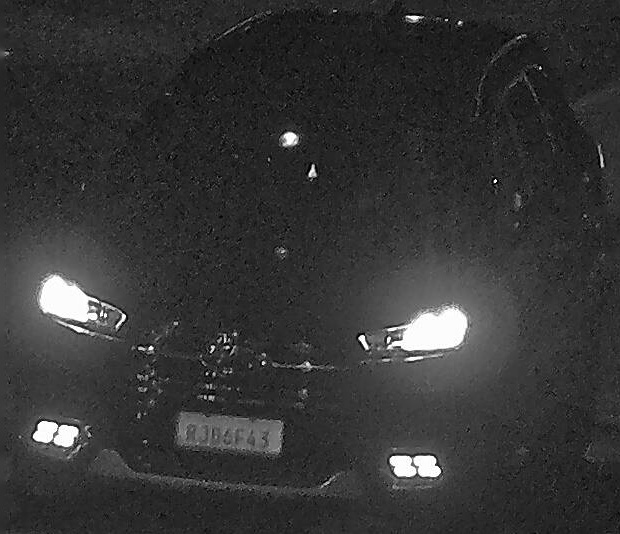}}
        \hspace{0.2mm}
        \subfloat[Light overexposure]{\includegraphics[height=12ex]{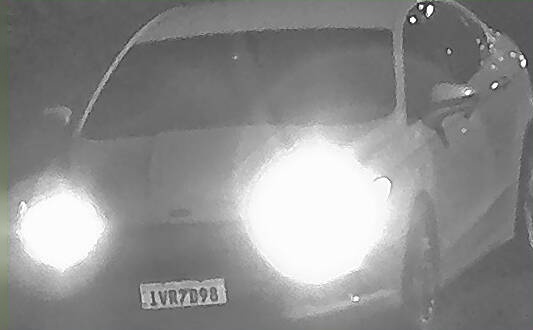}
        }
    }
    \caption{Examples of infrared images under varying conditions: (a) optimal visibility with enhanced perception, (b) reduced contrast affecting detail clarity, and (c) overexposure caused by vehicle headlights.}
    \label{fig:nighttime}
\end{figure}

The dataset spans a wide temporal range, including both daytime and nighttime conditions.
While timestamps are not available, images are categorized by the camera's capture mode. 
Nighttime images, primarily captured in infrared mode, account for $5{,}372$ images ($21.5$\%).
While infrared imaging improves visibility in low light, it also presents challenges, such as reduced contrast in fine details and potential overexposure from vehicle headlights (see \cref{fig:nighttime}).

\setcounter{figure}{0}
\begin{figure}[!htb]
    \captionsetup{captionskip=0pt}
    \centering
    \subfloat[Distribution of vehicle colors in the \dataset dataset.]{
        \includegraphics[width=0.9\linewidth]{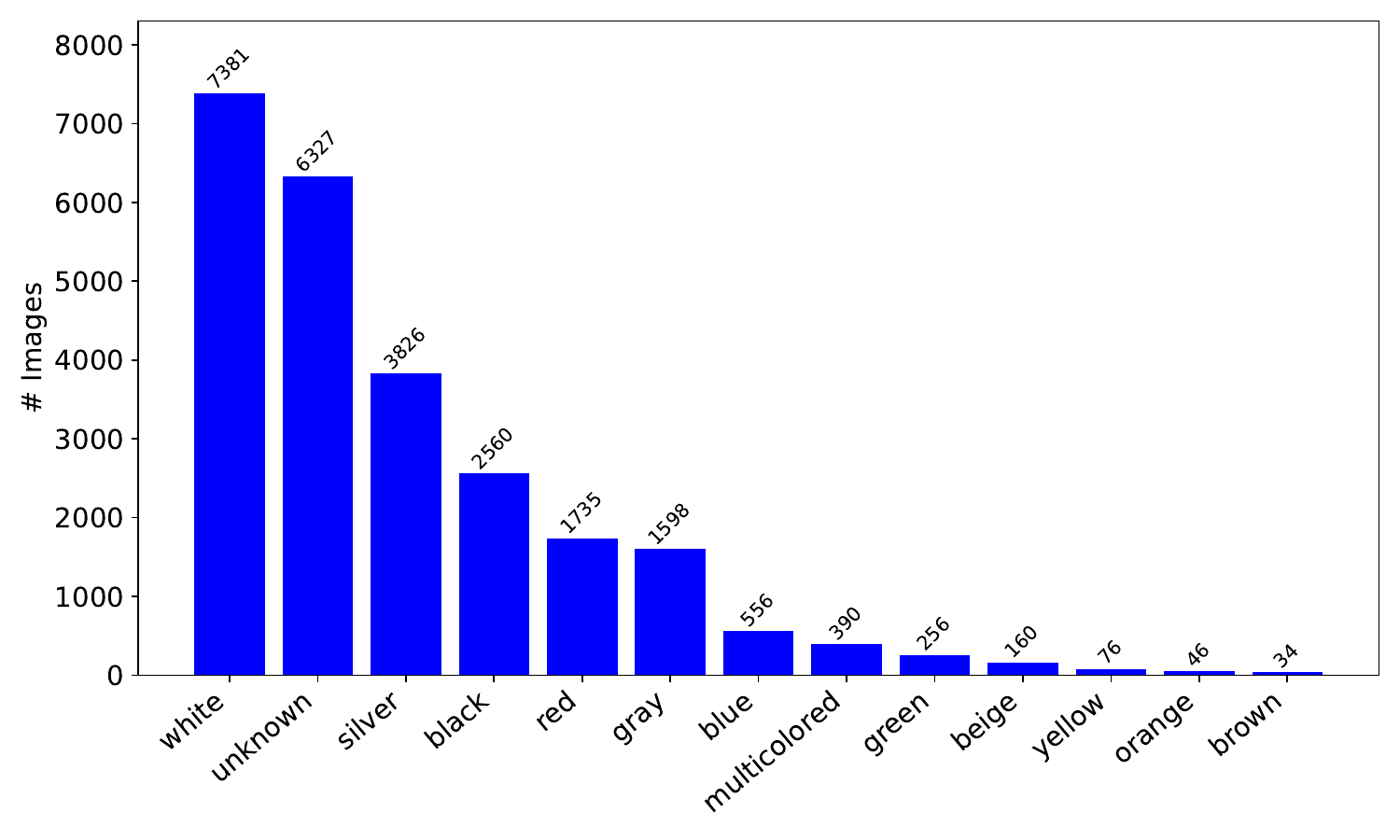}
    }
    \vspace{0.5cm}
    \subfloat[Distribution of vehicle makes in the \dataset dataset.]{
        \includegraphics[width=0.9\linewidth]{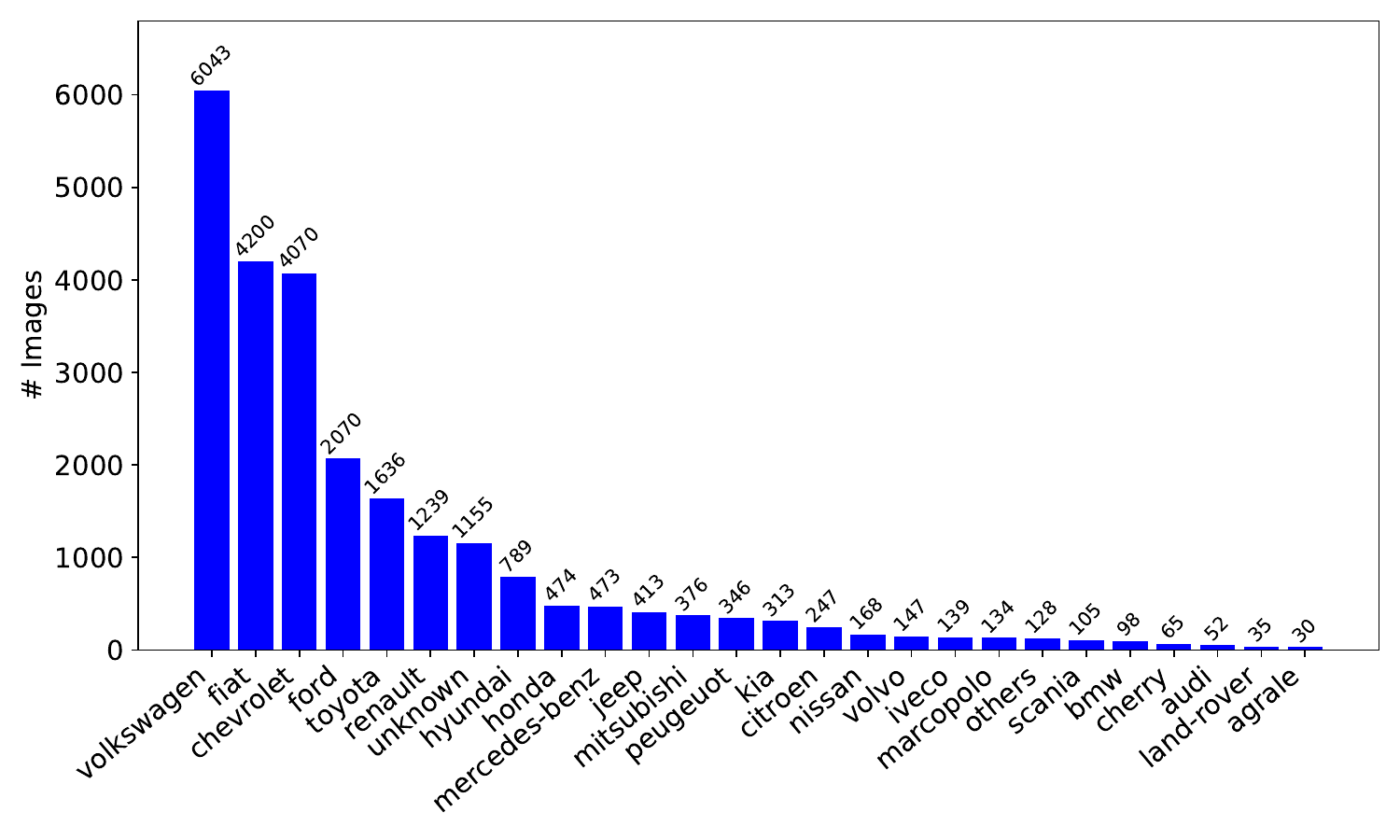}
    }
    \vspace{0.5cm}
    \subfloat[Distribution of the $30$ most common vehicle models in the \dataset dataset.]{
        \includegraphics[width=0.9\linewidth]{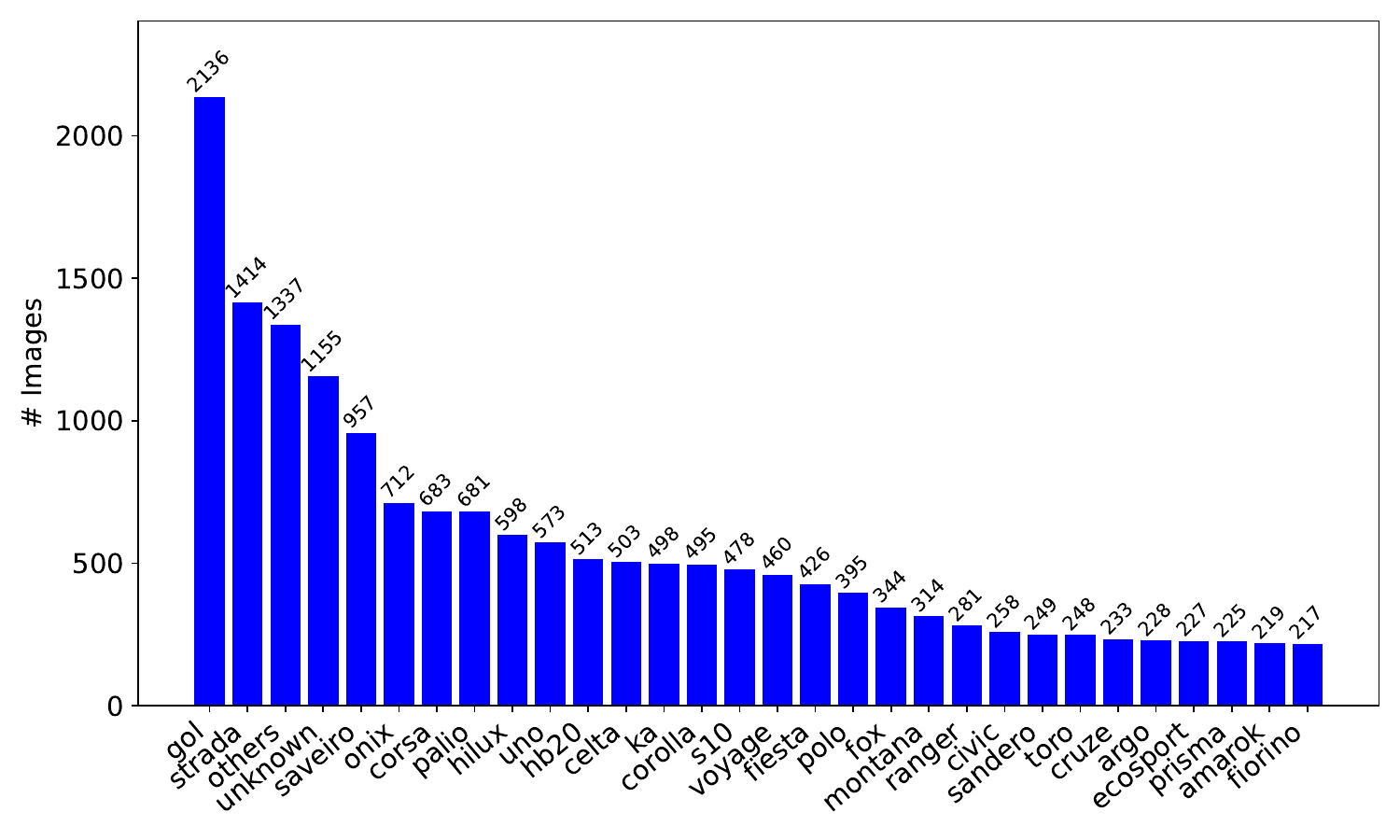}
    }
    \vspace{0.5cm}
    \subfloat[Distribution of vehicle types in the \dataset dataset.]{
        \includegraphics[width=0.9\linewidth]{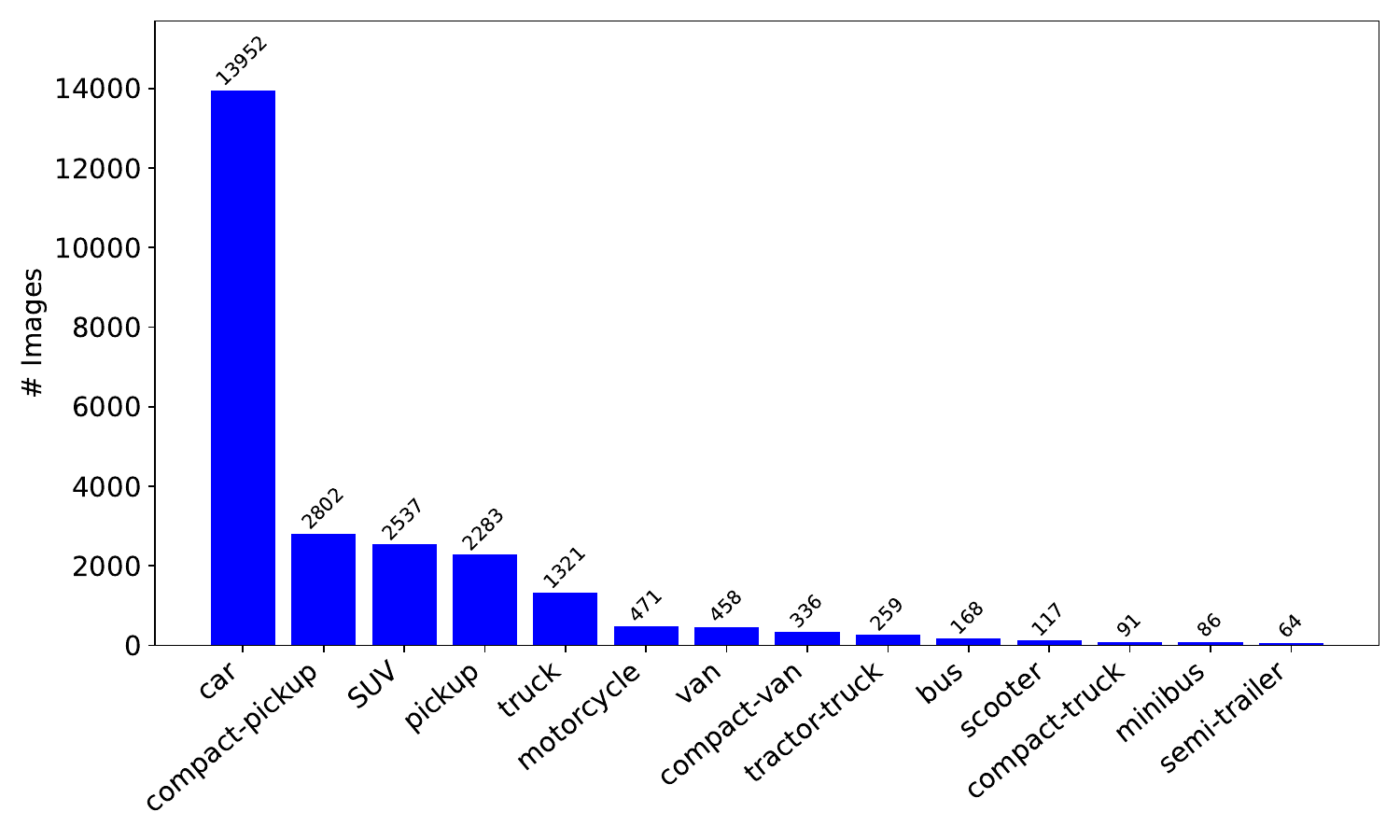}
    }
    \vspace{0.5mm}
    \caption{Distribution of vehicles across the attributes of color~(a), make~(b), model~(c) and type~(d) in the \dataset dataset. For better visualization, only the $30$ most common vehicle models are displayed in (c), representing $63.7$\% of the total images.
    }
    \label{fig:fgvc_class_distribution}
\end{figure}

Another factor contributing to the diversity of the dataset is the viewpoint of the vehicles.
\dataset captures multiple viewpoints, including frontal, rear, three-quarter, and high-angle perspectives. 
To standardize annotations, each image is categorized as either front or rear view based on the visibility of the \gls*{lp}. 
As a result, the dataset contains $13{,}842$ rear-view and $11{,}103$ frontal-view~images.

The dataset contains \major{$16{,}297$} \glspl*{lp} with two distinct layouts: Brazilian ($5{,}171$ \glspl*{lp}) and Mercosur ($11{,}126$ \glspl*{lp}). 
These \glspl*{lp} are captured under diverse conditions, including varying angles, resolutions, and levels of noise (see \cref{fig:sample_lps}). 
We remark that $0.2$\% of the \glspl*{lp} contain illegible characters. 
These images were retained due to their minimal impact on overall performance, their representation of real-world obstructions and degradations, and their non-interference with \gls*{fgvc} tasks. 
Importantly, accurate \gls*{lp} information was successfully retrieved even in cases involving occlusion (see \cref{subsec:annotation_process} for annotation details).

\setcounter{figure}{3}
\begin{figure}[!htb]
    \centering
    \captionsetup[subfigure]{captionskip=1.5pt,labelformat=empty}
    
    \resizebox{0.99\linewidth}{!}{
        \subfloat[\texttt{AYG5F32}]{\includegraphics[height=5.5ex]{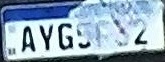}} \hspace{0.15mm}
        \subfloat[\texttt{MMI5A81}]{\includegraphics[height=5.5ex]{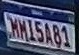}} \hspace{0.15mm}
        \subfloat[\texttt{RXS8I04}]{\includegraphics[height=5.5ex]{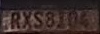}} \hspace{0.15mm}
        \subfloat[\texttt{AUG6H11}]{\includegraphics[height=5.5ex]{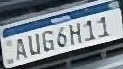}} \hspace{0.15mm}
        \subfloat[\texttt{LZT2H11}]{\includegraphics[height=5.5ex]{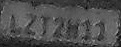}
        }
    }   

    \vspace{0.2cm}
    
    \resizebox{0.99\linewidth}{!}{
        \subfloat[\texttt{AYB5E38}]{\includegraphics[height=9ex]{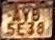}} \hspace{0.15mm}
        \subfloat[\texttt{SDV8A12}]{\includegraphics[height=9ex]{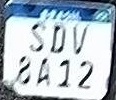}} \hspace{0.15mm}
        \subfloat[\texttt{AUN4331}]{\includegraphics[height=9ex]{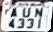}} \hspace{0.15mm}
        \subfloat[\texttt{AVX3J79}]{\includegraphics[height=9ex]{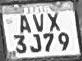}} \hspace{0.15mm}
        \subfloat[\texttt{AOP4214}]{\includegraphics[height=9ex]{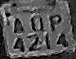}
        }
    }

    \vspace{0.3mm}
    
    \caption{Example of \glspl*{lp} cropped from images captured under diverse conditions, showcasing variations in resolution, perspective, and image quality. The corresponding annotated \gls*{lp} text is shown below each image.}
    \label{fig:sample_lps}
\end{figure}

\begin{figure}[!htb]
    \centering
    \includegraphics[width=0.98\columnwidth]{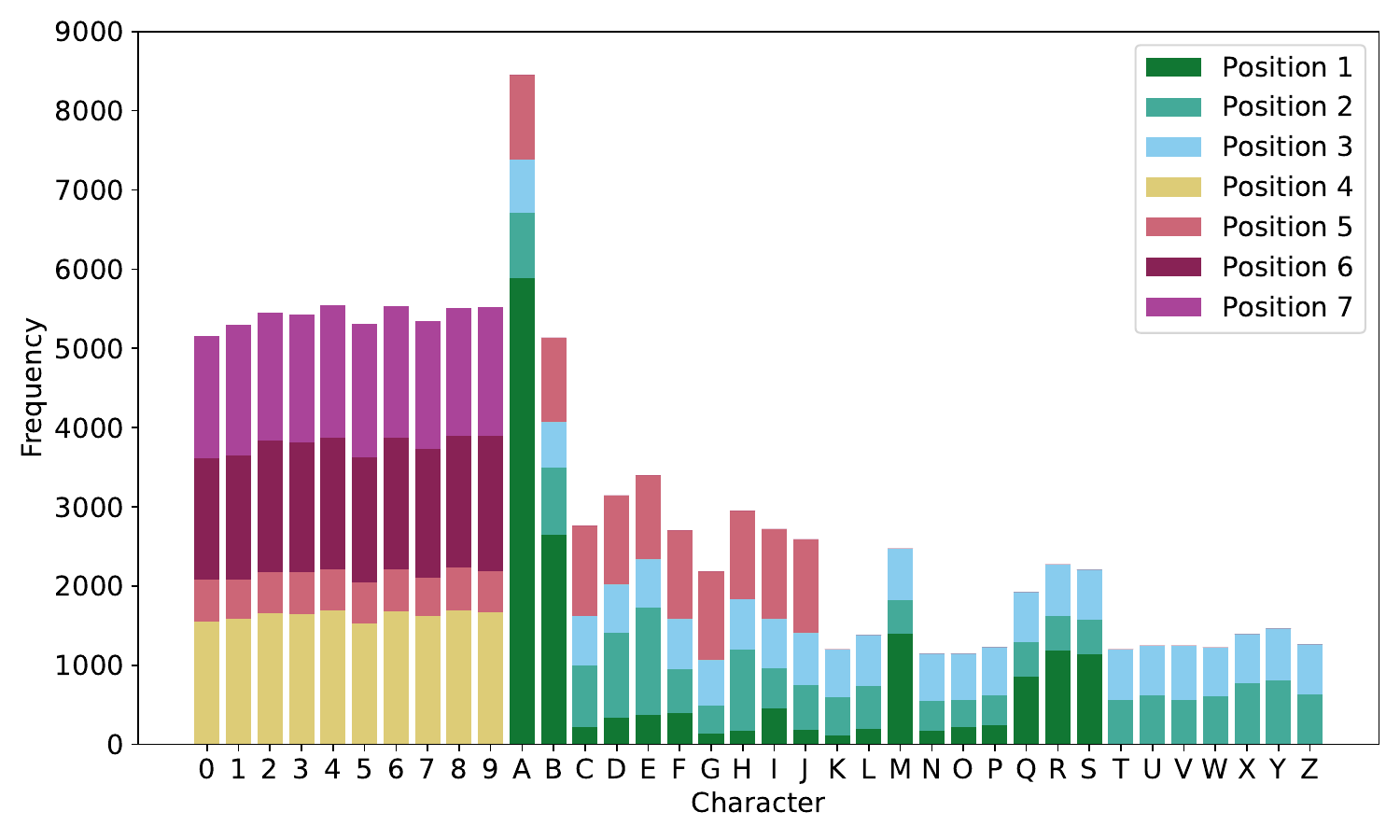}
    \caption{Distribution of \gls*{lp} character classes in the \dataset~dataset.
    }
    \label{fig:lp_char_frequency}
\end{figure}

Regarding \gls*{alpr}, the \dataset dataset includes annotations for both \gls*{lp} characters and corner coordinates.
\cref{fig:lp_char_frequency} shows the character distribution across the seven \gls*{lp} positions in the proposed dataset.
While digits are relatively evenly distributed, letters exhibit significant imbalances: LPs are more likely to start with certain letters, such as~``A'' and ``B''.
Character distribution in \gls*{alpr} datasets is inherently imbalanced due to region-specific \gls*{lp} allocation policies~\citep{goncalves2018realtime,laroca2022cross}.

With the dataset's key characteristics established, the following sections detail its creation process.
\cref{subsec:image_colection_preprocessing} elaborates on the image selection and preprocessing methods. 
\cref{subsec:annotation_process} describes the annotation methodology, including \gls*{lp} text labeling and \gls*{fgvc} attribute verification. 
\cref{subsec:splitting_protocol} defines the dataset splitting protocol to ensure reproducibility.
Finally, \cref{subsec:privacy_concerns} discusses the privacy safeguards implemented to address ethical~considerations.

\subsection{Image Collection and Preprocessing}
\label{subsec:image_colection_preprocessing}

Initially, $30{,}240$ images were collected from the Military Police of Paraná’s surveillance system. 
Each image was manually inspected to evaluate its suitability for the study, followed by a filtering process to eliminate samples that did not meet the research criteria.
A total of $1{,}253$ images were discarded due to factors such as extreme \gls*{lp} occlusions, severe image degradation, and poor vehicle framing.
Furthermore, the images were grouped based on \gls*{lp} information, and highly similar samples --~such as those with similar viewpoints and lighting conditions~-- were removed, resulting in the exclusion of an additional $4{,}042$~images.

As the police system collects images from various surveillance sources, the dataset exhibited inconsistencies in format.
Some images were already cropped around vehicles, while others included significant background content.
To standardize the dataset, the YOLOv11 model~\citep{yolov11} was employed for vehicle detection and precise cropping.
This model was selected due to its strong detection performance and its widespread adoption in both academic and industrial applications~\citep{nyi2025two,he2024research,khanam2024yolov11}.

A manual review was carried out to ensure proper vehicle cropping, especially in images containing multiple detected vehicles.
Manual intervention was also necessary in cases where the model failed to detect a vehicle or produced inaccurate results.
This was particularly important in challenging scenarios, such as motorcycles captured in low-light conditions or vehicles appearing at a~distance.

\begin{figure}[!htb]
    \centering
    \captionsetup[subfigure]{captionskip=1.5pt,font=scriptsize}
        \subfloat[Original]{\includegraphics[height=14ex]{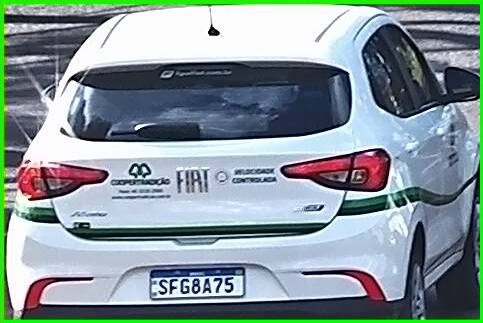}}
        \hspace{0.2mm}
        \subfloat[Processed]{\includegraphics[height=14ex]{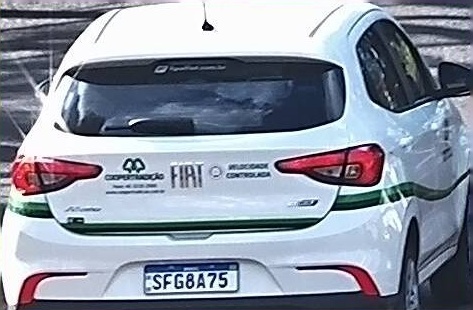}}
    \caption{Example of border standardization. (a)~original image with a green border; (b)~result after removing a $5$-pixel margin from all sides.}
    \label{fig:green_border}
\end{figure}

Additional standardization was performed to address the presence of green borders found in some pre-cropped images.
To ensure visual consistency, a $5$-pixel border was removed from all sides of each image.
A manual review was then conducted to verify that no \glspl*{lp} were significantly affected or occluded by this adjustment.
\cref{fig:green_border} illustrates an example of this issue and the resulting image after~processing.

Finally, it is important to highlight that the images were not resized to uniform dimensions. 
As a result, the dataset retains a variety of image sizes, with widths ranging from $89$ to $2{,}110$ pixels and heights from $135$ to $1{,}408$ pixels. 
This approach prevents distortions that could affect recognition tasks, allowing resizing to be handled as needed for specific models and~applications.

\subsection{Annotation Process}
\label{subsec:annotation_process}

\gls*{lp} text annotations were initially performed manually and subsequently used to automate the retrieval of \gls*{fgvc} attributes via the \gls*{senatran} database. 
The retrieved vehicle information was then manually verified against the vehicle's visual appearance to ensure consistency.
In cases of discrepancies, the \gls*{lp} text was re-annotated manually, followed by another round of automated retrieval and verification of the \gls*{fgvc}~attributes.

The use of manually annotated \glspl*{lp} during the \gls*{fgvc} annotation process carries an important implication: the \glspl*{lp} in the dataset are human-recognizable. 
While the dataset captures an unconstrained surveillance scenario for \gls*{fgvc} tasks, the \gls*{alpr} context had to be restricted to ensure \gls*{lp} readability.
This introduces a limitation in fully replicating real-world conditions, as it substantially reduces the presence of challenging cases typically faced by \gls*{alpr}~systems.

However, this limitation represents a necessary step toward advancing \gls*{fgvc} research and its integration with \gls*{alpr}. 
To date, no existing study has jointly addressed vehicle color, make, model, and type recognition, nor explored their combined use with \gls*{lp} recognition (as detailed in \cref{sec:RelatedWork}).
The primary goal of \dataset is to establish a foundation for future research in this direction. 
We anticipate that future datasets will build upon and enhance the scenarios and discussions presented here, ultimately contributing to the progress of both \gls*{fgvc} and \gls*{alpr}~fields.

To better support \gls*{fgvc}-related tasks, the color, make, and model annotations in the \dataset dataset were refined to maximize its utility across all attributes.
These adjustments aimed to include previously overlooked scenarios, enabling a more comprehensive evaluation of their impact on recognition performance and better reflecting the challenges of real-world \gls*{its} applications.

\begin{figure}[!htb]
    \centering
    \captionsetup[subfigure]{captionskip=1.5pt,font=scriptsize}
    \subfloat[]{\includegraphics[height=15ex]{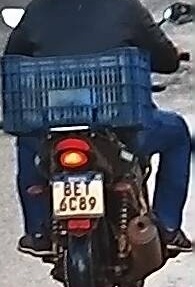}\label{subfig:unknown_motorcycle}}
    \hspace{0.2mm}
    \subfloat[]{\includegraphics[height=15ex]{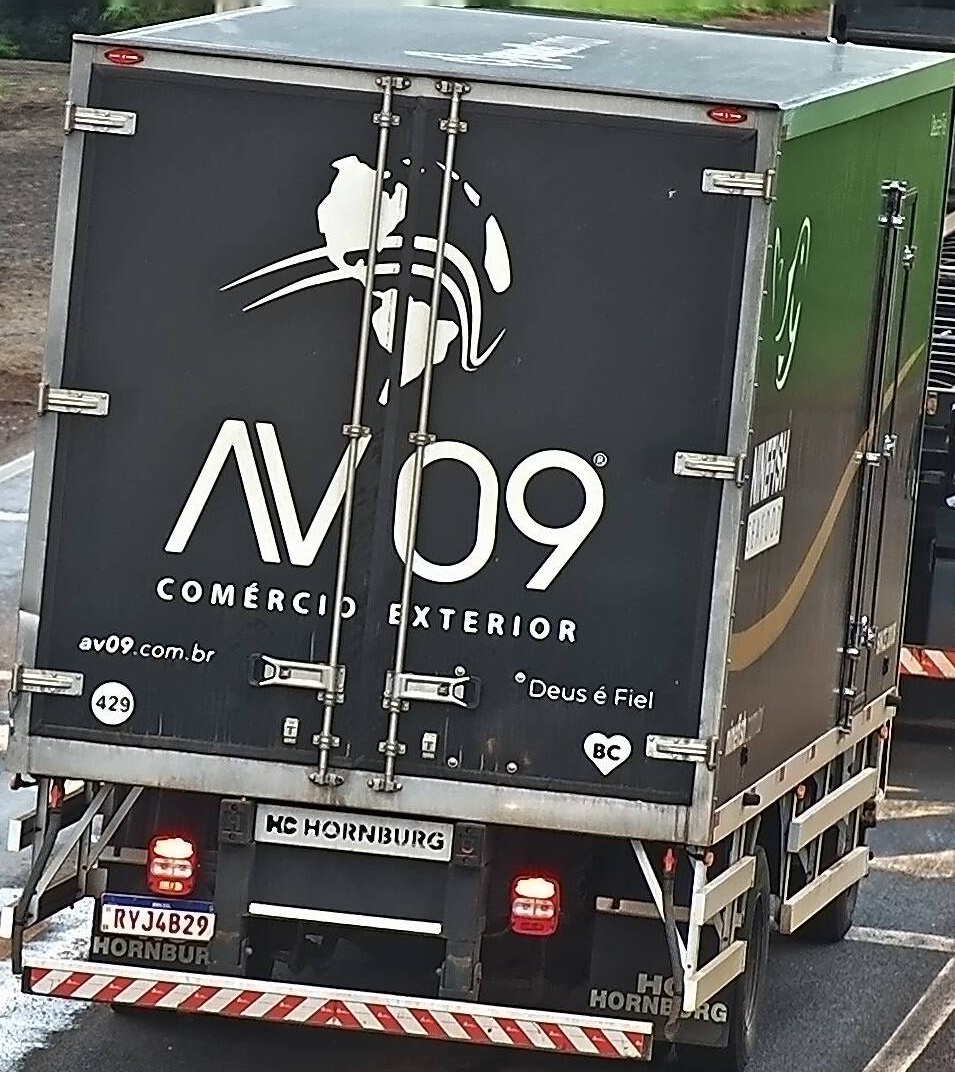}\label{subfig:unknown_white_truck}}
    \hspace{0.2mm}
    \subfloat[]{\includegraphics[height=15ex]{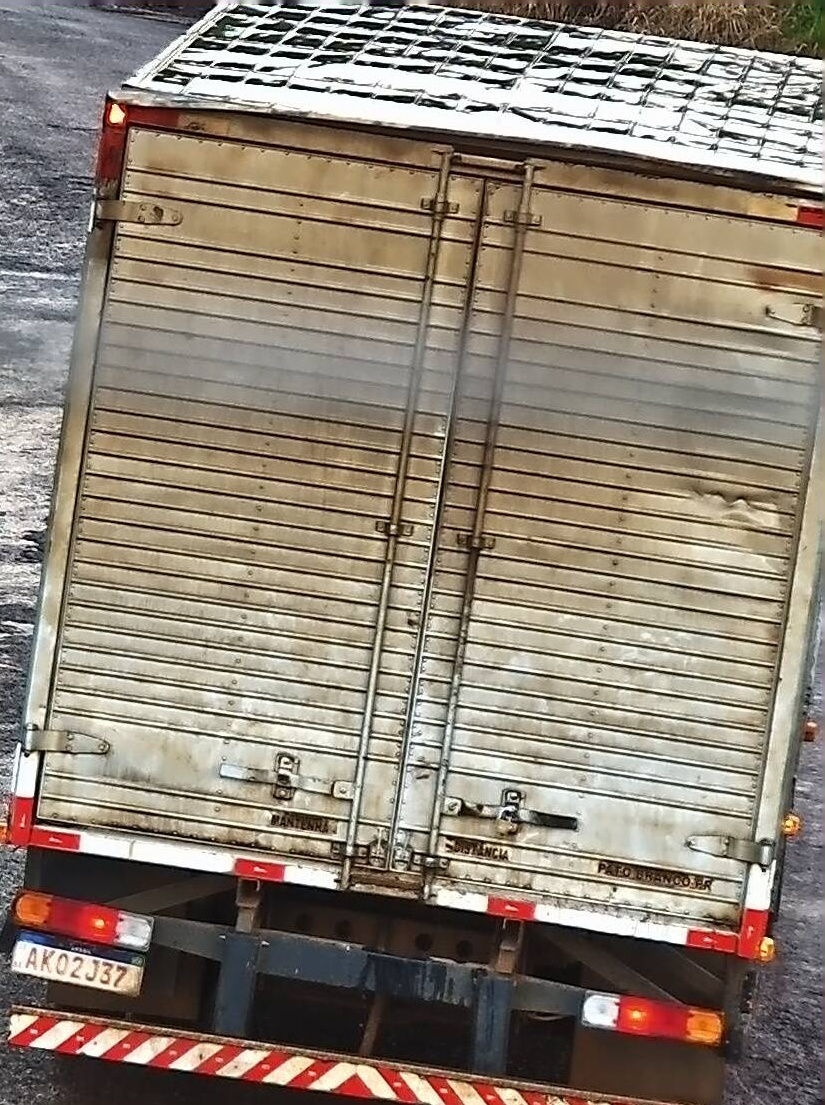}\label{subfig:unknown_red_truck}}

    \vspace{0.3mm}
    
    \caption{Examples of vehicles assigned to the ``unknown'' class for color, make, and model annotations: (a)~a red Yamaha motorcycle; (b)~a white Volkswagen truck; (c)~a red Mercedes-Benz truck. In all cases, the main body of the vehicle is obstructed, rendering the identification of color, make, and model~\major{impossible}.}
    \label{fig:unknown_scenarios}
\end{figure}

Infrared images were grouped into a dedicated color class due to their inherent lack of color data. 
Vehicles officially categorized or visually similar to multicolored -- a non-literal translation of \textit{fantasia} in Portuguese, used when no predominant color is identifiable -- were also retained in their own color class. 
Motorcycles and scooters had color, make, and model attributes reassigned to an ``unknown'' class due to their underexplored nature in \gls*{fgvc} literature and the limited visual information from these images~(\cref{subfig:unknown_motorcycle}). 
The same strategy was applied to rear-view truck-like vehicles, where cargo compartments frequently obstruct the main body, making accurate \gls*{fgvc} \major{impossible}~(\cref{subfig:unknown_white_truck,subfig:unknown_red_truck}).

Classes with fewer than $25$ samples were adjusted to reduce extreme class imbalance. 
Underrepresented colors were merged into visually similar classes (e.g., purple into blue, garnet into red, gold into beige) while underrepresented make/model were consolidated into a generic ``others'' class.
Additionally, vehicle models from different production years were grouped into a single class, in contrast to related datasets~\citep{yang2015compcars,kuhn2021brcars}. 
This decision was made to avoid excessive fragmentation, ensuring both class balance and experimental~reliability.

Corner annotations were obtained for each \gls*{lp} using a two-stage approach.
We use a YOLOv11 model~\citep{yolov11} to detect the \gls*{lp} regions within the input image, and then CDCC-NET~\citep{laroca2021towards} regressed the corner coordinate values.
Both models were fine-tuned on popular public datasets collected from multiple regions, such as AOLP~\citep{hsu2013aolp}, UFPR-ALPR~\citep{laroca2018robust}, CLPD~\citep{zhang2021robust}, among others.
A matching algorithm was used to handle multiple detections within an image, assigning a match if all \gls*{lp} corners fell within the vehicle's bounding box. 
In cases where multiple plates were visible (as in \cref{fig:multiple_vehicles}), manual identification was performed.
Manual corrections were applied when the YOLOv11 model misidentified non-\gls*{lp} textual regions.

Additional annotations --~such as vehicle viewpoint and camera capture mode~-- were manually labeled and carefully verified to improve the overall quality of the~dataset.

\subsection{Splitting Protocol}
\label{subsec:splitting_protocol}

To ensure robust and unbiased model evaluation on the \dataset dataset, we developed a custom $5$-fold evaluation protocol. 
This methodology is specifically designed to prevent data leakage and mitigate partitioning bias, thereby providing a more reliable estimate of a model's true generalization performance.

The protocol begins by partitioning the entire dataset into five non-overlapping folds, a process governed by two critical constraints. 
The first constraint is the prevention of data leakage, an issue that can lead to overly optimistic performance estimates~\citep{laroca2023do}; this is achieved by confining all images of the same vehicle (i.e. with the same \gls*{lp}) to a single fold. 
The second constraint is multi-attribute stratification, where partitioning is performed simultaneously across all four vehicle attributes -- color, make, model, and type -- to ensure each fold preserves the class distribution of the original dataset as close as possible.

Following that, we generated ten distinct $3$:$1$:$1$ train-validation-test splits from the five folds.
Our protocol is designed to deterministically use each of the $C(5, 3) = 10$ unique combinations of three folds as the training set exactly once.
To achieve this, each of the five folds (indexed $0$ to $4$) is used as the test set exactly twice.
The two validation sets for that test set are then assigned using the rules $val_1 = (test + 1) \pmod 5$, and $val_2 = (test + 2) \pmod 5$.
The training set for each split is subsequently formed by the three remaining folds.
The specific files defining all train-validation-test splits are publicly available to ensure reproducibility.

\subsection{Privacy concerns}
\label{subsec:privacy_concerns}

To comply with ethical and legal guidelines, privacy-sensitive elements were addressed. 
While \glspl*{lp} do not constitute personal data in Brazil -- since they cannot be directly linked to vehicle owners -- some images contain identifiable faces of drivers or pedestrians. 
To mitigate this, the RetinaFace model~\citep{deng2020retinaface} was employed to detect facial regions for further blurring. 
After automated processing, a manual verification step was conducted, allowing for minor corrections to ensure the quality and effectiveness of the anonymization process.

%% file: 4-analysis.tex
\section{Comparative Analysis}
\label{sec:ComparativeAnalysis}

This section provides a two-stage comparative analysis to highlight the contributions of the \dataset dataset. 
First, a qualitative analysis (\cref{subsec:qualitative_analysis}) reveals that preceding benchmarks are often limited and fail to capture real-world diversity, rendering them less complex. 
Subsequently, a quantitative analysis (\cref{subsec:quantitative_analysis}) provides empirical evidence for this claim. 

\subsection{Qualitative Analysis}
\label{subsec:qualitative_analysis}

\cref{tab:rw_datasets} provides a structured comparison of related datasets. 
The total number of images and unique vehicle identities are also included. 
When details were not provided in the original studies, they are marked as unknown (unk.). 
The number of classes is omitted to prevent inconsistencies, as different datasets may define class labels in varying ways. 
For instance, some datasets merge make and model information into a single class, while others treat them hierarchically.

\begin{table}[!htb]
    \centering
    \caption{Comparison between \dataset (proposed in this work) and related~datasets.}
    \label{tab:rw_datasets}
    \resizebox{0.998\columnwidth}{!}{
        \begin{tabular}{lrrllcccc}
            \toprule
            \multirow{2}{*}{Dataset}
            &\multirow{2}{*}{Images}
            &\multirow{2}{*}{Vehicles}
            &\multirow{2}{*}{Source}
            &\multirow{2}{*}{Viewpoint}
            &\multicolumn{4}{c}{Annotations} \\
            \cmidrule{6-9}
            & & & & &Color &Make/Model &Type &ALPR \\
            \midrule
                \cite{ferryman1995generic} &$176^{\phantom{\ast}}$ &unk. &Field &Frontal/Rear &- &- &\checkmark  &- \\
                \cite{jollyvehicle1996} &$393^{\phantom{\ast}}$ &unk. &Field &n/a &- &- &\checkmark  &- \\
                \cite{lai2001vehicle} &unk.$^{\phantom{\ast}}$ &unk. &Field &Rear &- &- &\checkmark  &- \\
                \cite{wu2001method} &$800^{\phantom{\ast}}$ &unk. &Field &Frontal &- &- &\checkmark  &- \\
                \cite{ma2005edge} &unk.$^{\phantom{\ast}}$ &unk. &Field &Frontal &- &- &\checkmark  &- \\
                \cite{baek2007vehicle} &$500^{\phantom{\ast}}$ &unk. &Field &Frontal &\checkmark  &- &- &- \\
                \cite{dule2010convenient} &$1{,}960^{\phantom{\ast}}$ &unk. &Field &Frontal &\checkmark  &- &- &- \\
                AOLP~\citep{hsu2013aolp} &$2{,}049^{\phantom{\ast}}$ &$1{,}286$ &Field &Frontal/Rear &- &- &- &\checkmark  \\
                Stanford Cars-196~\citep{krause2013collecting} &$16{,}185^{\phantom{\ast}}$ &unk. &Web &Frontal/Rear &- &\checkmark  &- &- \\
                \cite{chen2014vehicle} &$15{,}601^{\phantom{\ast}}$ &unk. &Field &Frontal &\checkmark  &- &- &- \\
                BIT-Vehicle~\citep{dong2015vehicle} &$9{,}850^{\phantom{\ast}}$ &unk. &Field &Frontal &- &- &\checkmark  &- \\
                CompCars-SV~\citep{yang2015compcars}  &$44{,}481^{\phantom{\ast}}$ &unk. &Field &Frontal &\checkmark  &\checkmark  &- &- \\
                BoxCars~\citep{sochor2016boxcars} &$63{,}750^{\phantom{\ast}}$ &$21{,}250$ &Field &Frontal/Rear &- &\checkmark  &- &- \\
                SSIG-ALPR~\citep{goncalves2018realtime} &$2{,}000^{\phantom{\ast}}$ &$815$ &Field &Frontal/Rear &- &- &- &\checkmark  \\
                PKU~\citep{yuan2017robust} &$3{,}977^{\phantom{\ast}}$ &$1{,}933$ &Field &Frontal &- &- &- &\checkmark  \\
                SYSU-Vehicle~\citep{hu2017location-aware} &$5{,}000^{\phantom{\ast}}$ &unk. &Web &Frontal/Rear &- &- &\checkmark  &- \\
                CCPD~\citep{xu2018towards} &$250{,}000^{\phantom{\ast}}$ &unk. &Field &Frontal/Rear &- &- &- &\checkmark  \\
                UFPR-ALPR~\citep{laroca2018robust} &$4{,}500^{\phantom{\ast}}$ &$150$ &Field &Frontal/Rear &- &- &- &\checkmark  \\
                \cite{shvai2020accurate} &$73{,}638^{\phantom{\ast}}$ &unk. &Field &Frontal &- &- &\checkmark  &- \\
                MPF-Cars~\citep{wang2020multipath} &$335{,}011^{\phantom{\ast}}$ &$71{,}305$ &Field &Frontal &- &\checkmark  &- &- \\
                BR-Cars~\citep{kuhn2021brcars} &$300{,}325^{\phantom{\ast}}$ &$52{,}000$ &Web &Frontal/Rear &- &\checkmark  &- &- \\
                Vehicle-Rear~\citep{oliveira2021vehicle} &$26{,}160^{\ast}$ &$2{,}966$ &Field &Rear &\checkmark  &\checkmark  &- &\checkmark  \\
                \cite{wang2021transformer} &$32{,}220^{\phantom{\ast}}$ &unk. &Field &Rear &\checkmark  &- &- &- \\
                RodoSol-ALPR~\citep{laroca2022cross} &$20{,}000^{\phantom{\ast}}$ &$12{,}785$ &Field &Frontal/Rear$^{\dagger}$ &- &- &- &\checkmark  \\
                DeepCar~5.0~\citep{amirkhani2023deepcar} &$40{,}185^{\phantom{\ast}}$ &unk. &Web &Frontal &- &\checkmark  &- &- \\
                Vehicle Color-24~\cite{hu2023vehicle} &$31{,}232^{\phantom{\ast}}$ &unk. &Field &Frontal &\checkmark  &- &- &- \\
                \cite{basak2024vehicle} &$7{,}242^{\phantom{\ast}}$ &unk. &Field &Frontal/Rear &- &- &\checkmark  &- \\
                \cite{luo2024dense_tnt} &$57{,}984^{\phantom{\ast}}$ &unk. &Field &n/a &- &- &\checkmark  &- \\
                UFPR-VCR~\citep{lima2024toward} &$10{,}039^{\phantom{\ast}}$ &$9{,}502$ &Field &Frontal/Rear &\checkmark  &- &- &- \\
                \midrule
                \textbf{\dataset (ours)} &$24{,}945^{\phantom{\ast}}$ &$16{,}297$ &Field &Frontal/Rear &\checkmark  &\checkmark  &\checkmark  &\checkmark  \\
            \bottomrule
            \multicolumn{5}{l}{\rule{-3pt}{2.1ex}$^{\ast}$\hspace{0.2mm}Counting reported for Vehicle-Rear vehicle frames.} \\
            \multicolumn{5}{l}{\rule{-3pt}{2.1ex}$^{\dagger}$\hspace{0.2mm}Only motorcycle images are rear-view.} \\
        \end{tabular}
    }
\end{table}

\REWRITTEN{
Following the \gls*{fgvc} literature, we also categorize datasets based on their image sources. 
Field-sourced datasets are derived from surveillance or traffic monitoring systems, capturing vehicles in their natural operational conditions. 
In contrast, web-sourced datasets comprise images gathered online from advertisements and curated photographic scenes. 
Although web-sourced datasets contribute to \gls*{fgvc} research, they often fail to reflect real-world surveillance conditions.
}

\REWRITTEN{
Datasets are further categorized based on the images' viewpoint. 
Frontal and rear viewpoints indicate which plate is in view.
Other perspectives are labeled as not applicable (n/a) due to the absence of a visible plate, which are needed for our \gls*{alpr} experiments and its integration with \gls*{fgvc}.
}

\REWRITTEN{
As \cref{tab:rw_datasets} illustrates, most datasets specialize in either \gls*{fgvc} or \gls*{alpr}, and \gls*{fgvc} benchmarks typically treat attributes like color, type, and make/model independently. 
The only exceptions that annotate all these attributes are Vehicle-Rear and \dataset. 
However, Vehicle-Rear is limited to $3{,}000$ unique vehicles from a single viewpoint and was not used for \gls*{fgvc} in its original study. 
In contrast, \dataset provides a more diverse collection of over $16{,}000$ distinct vehicles captured from multiple viewpoints and under varied real-world conditions.
}

\begin{figure}[!t]
    \centering
    \captionsetup[subfigure]{labelformat=empty,captionskip=2.5pt}
    \resizebox{0.9\linewidth}{!}{
        \subfloat[]{
                \centering              
                \includegraphics[height=12.8ex]{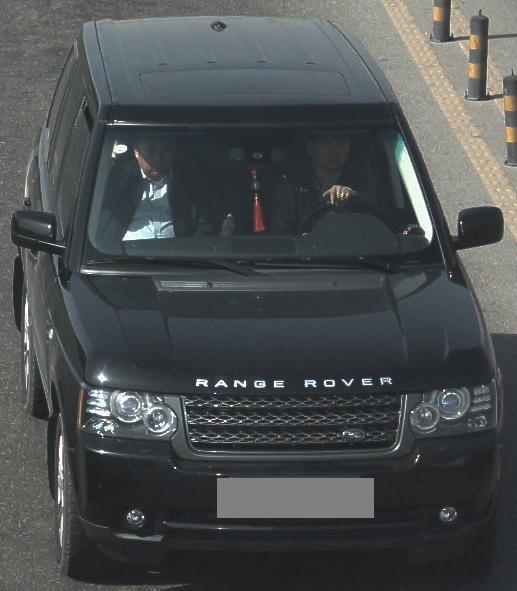}
                \includegraphics[height=12.8ex]{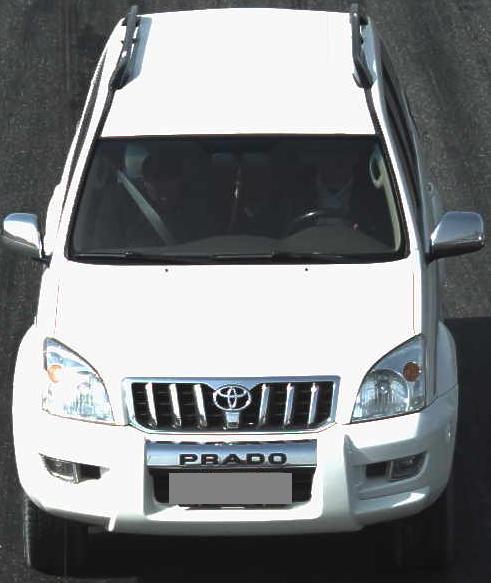}
                \includegraphics[height=12.8ex]{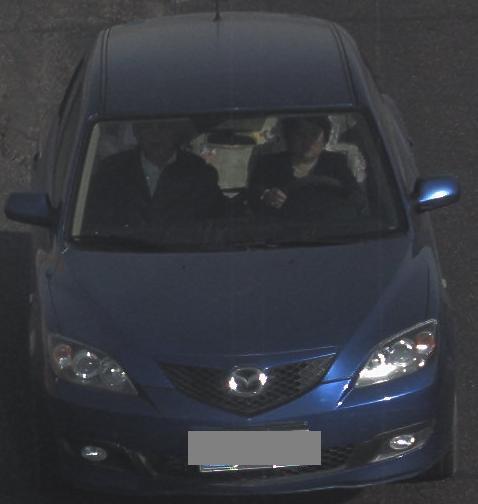}
                \includegraphics[height=12.8ex]{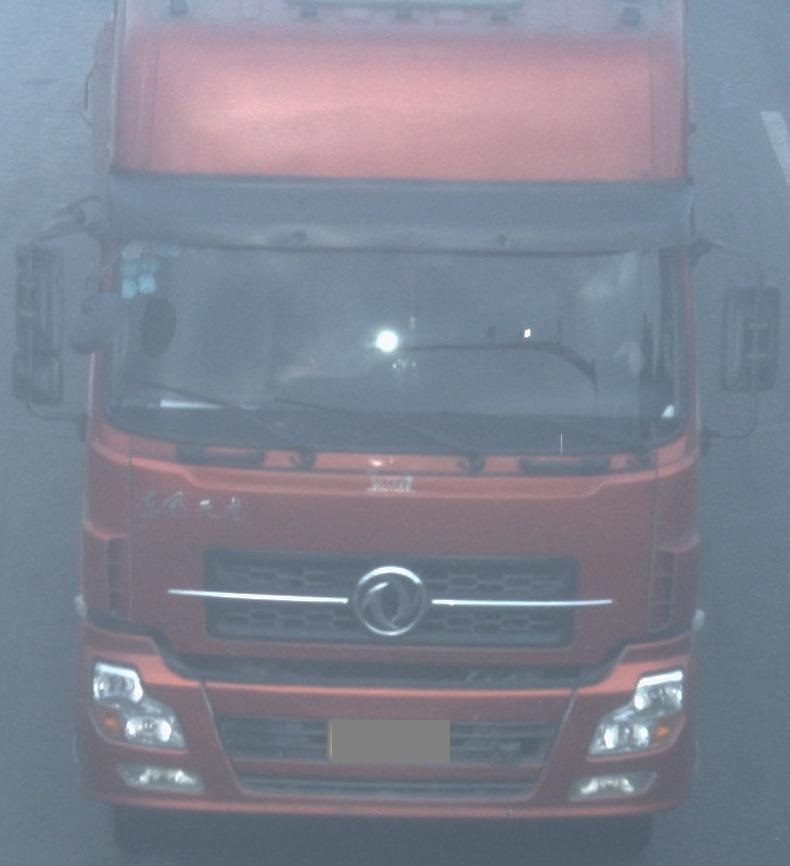}
        }
    }

        \vspace{-3.75mm}

    \resizebox{0.9\linewidth}{!}{
        \subfloat[(a)~Images from the dataset proposed by \cite{chen2014vehicle}.]{
                \includegraphics[height=11ex]{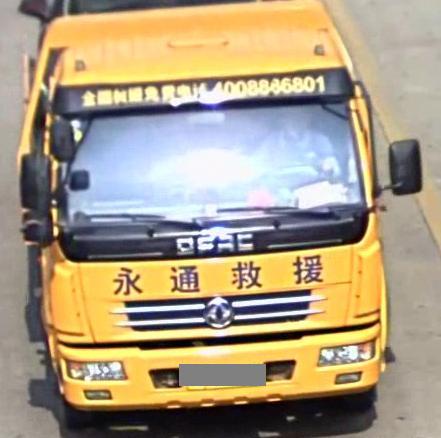}
                \includegraphics[height=11ex]{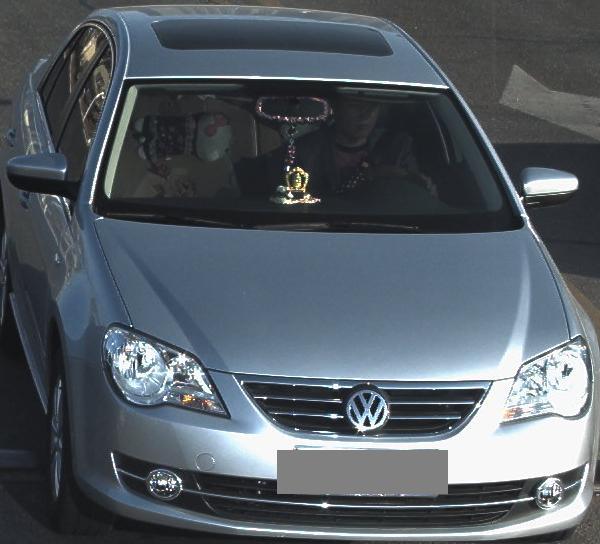}
                \includegraphics[height=11ex]{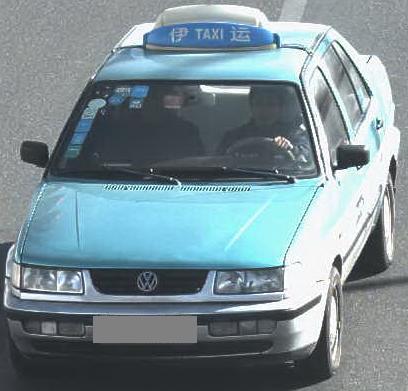}
                \includegraphics[height=11ex]{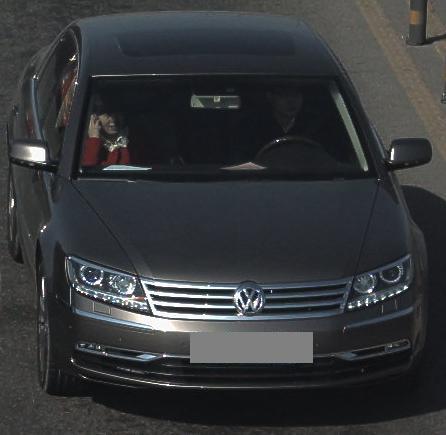}
        }
    }
    
    \vspace{0.25cm} 
    
    \resizebox{0.9\linewidth}{!}{
        \subfloat[]{
                \centering
                
                \includegraphics[height=12.75ex]{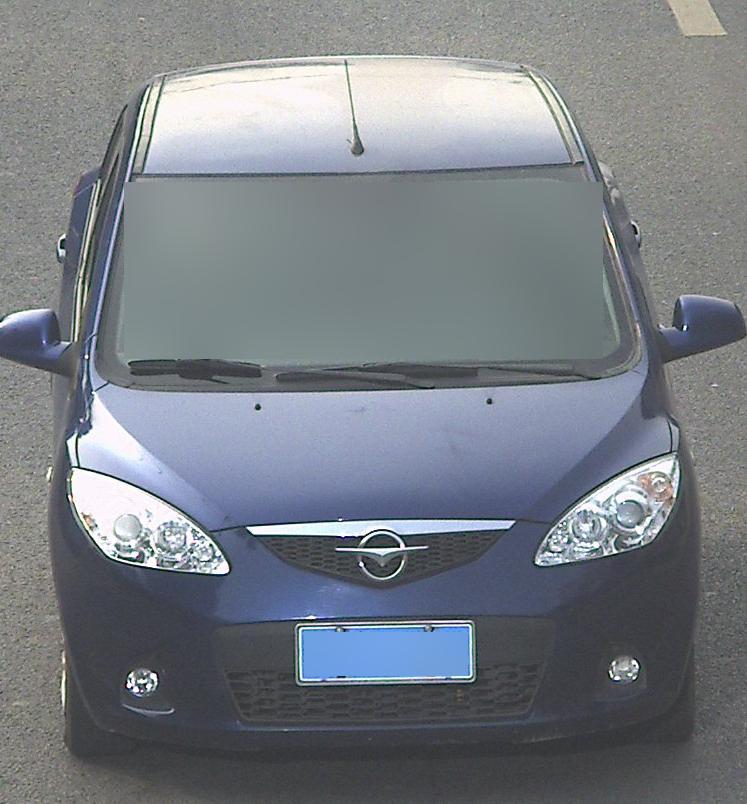}
                \includegraphics[height=12.75ex]{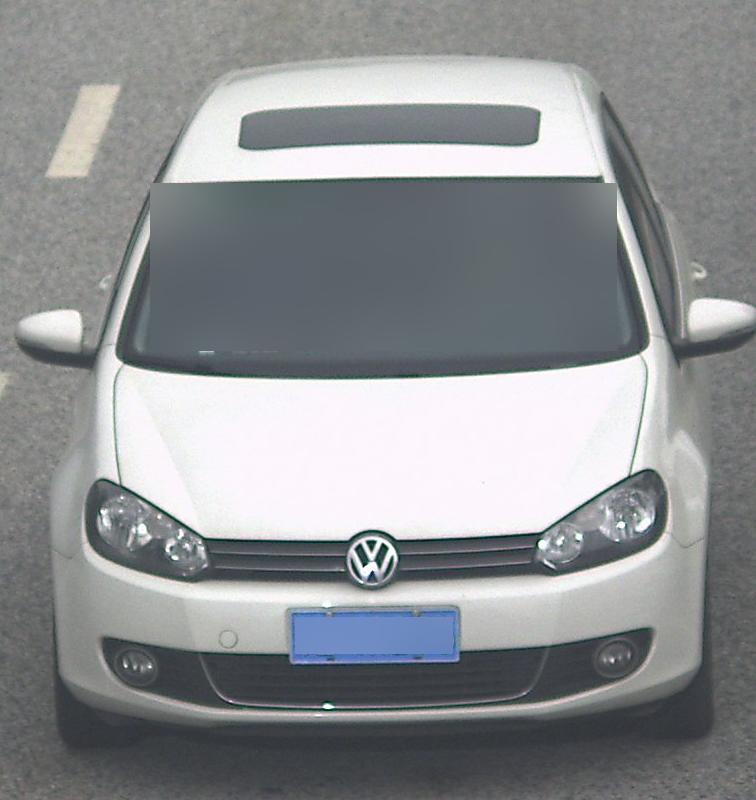}
                \includegraphics[height=12.75ex]{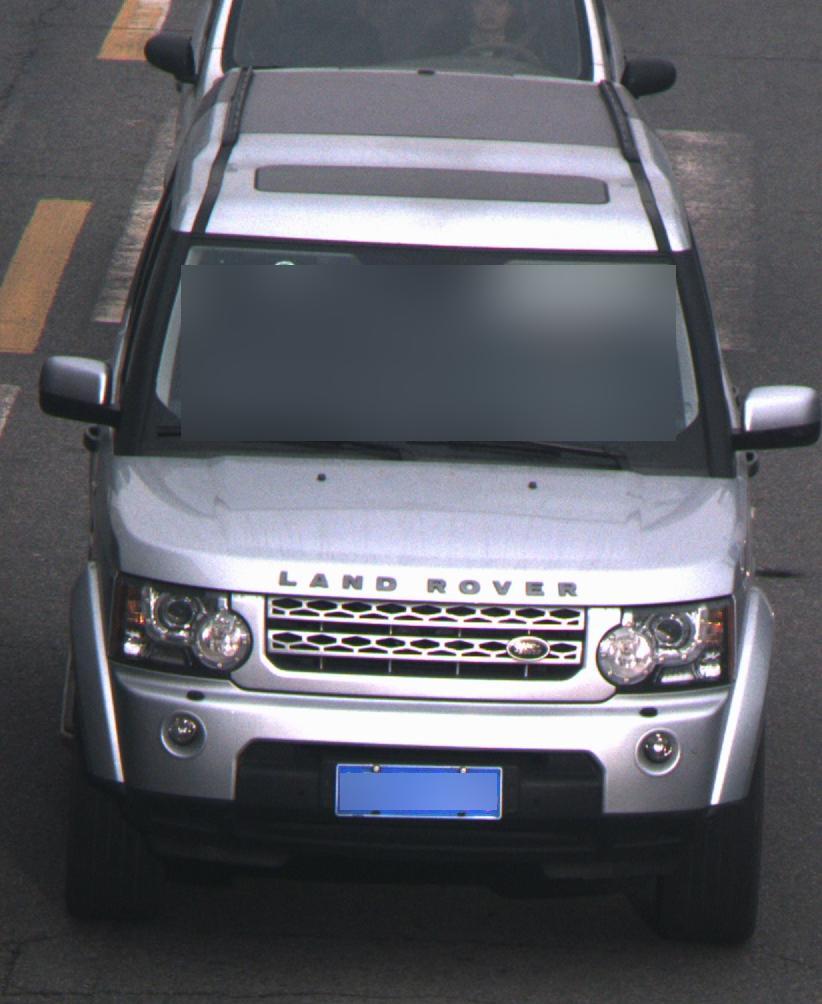}
                \includegraphics[height=12.75ex]{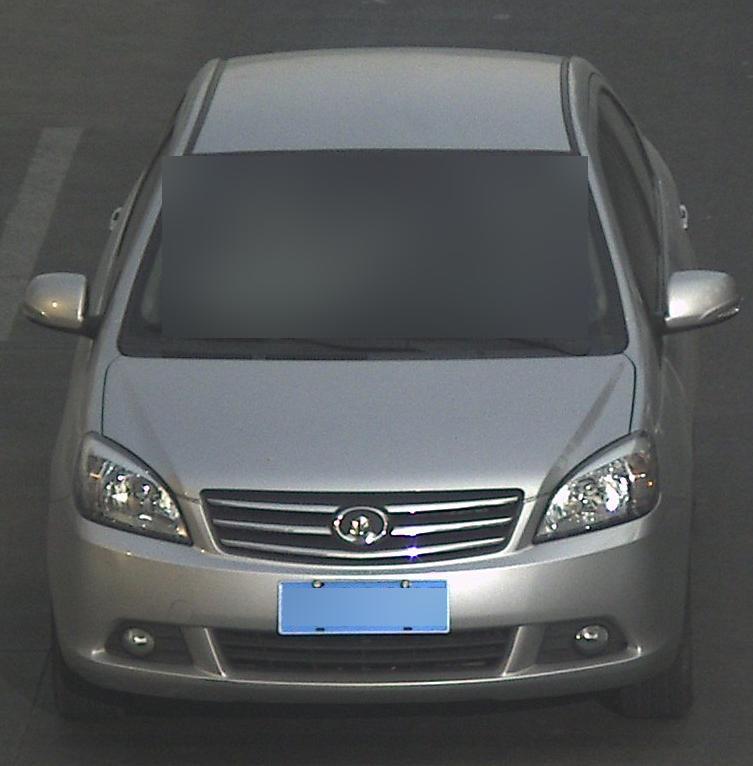}
        }
    }
                
    \vspace{-3.75mm}
                
    \resizebox{0.9\linewidth}{!}{
        \subfloat[(b)~Images from CompCars-SV, proposed by \cite{yang2015compcars}.]{
                \includegraphics[height=12.75ex]{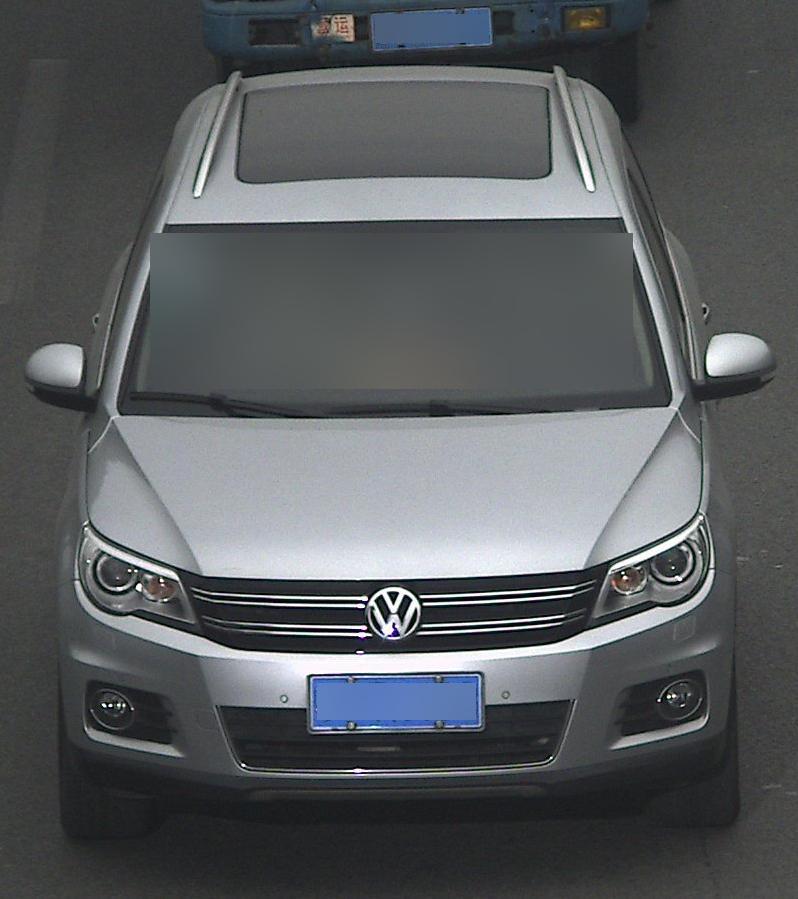}
                 \includegraphics[height=12.75ex]{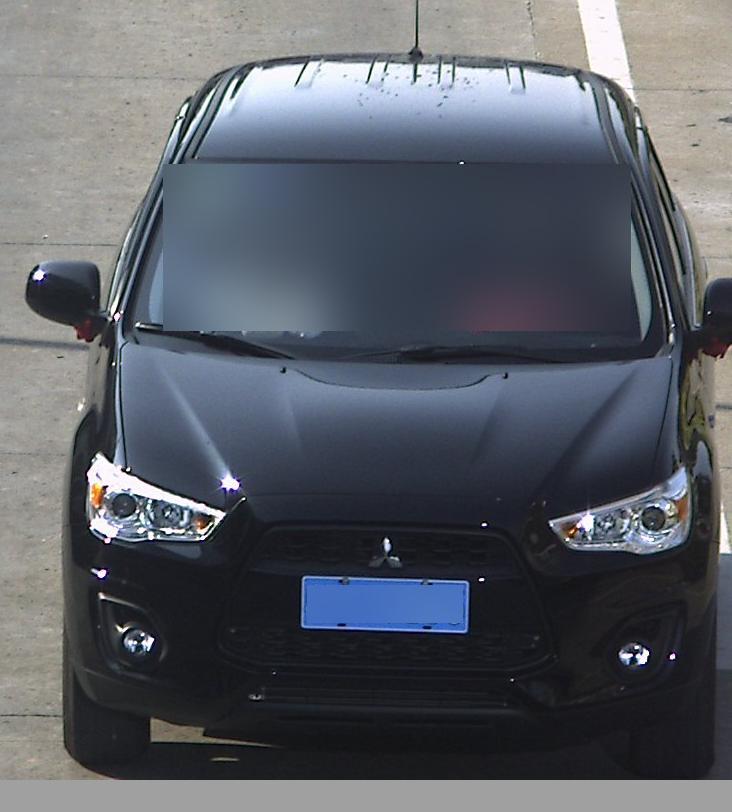}
                \includegraphics[height=12.75ex]{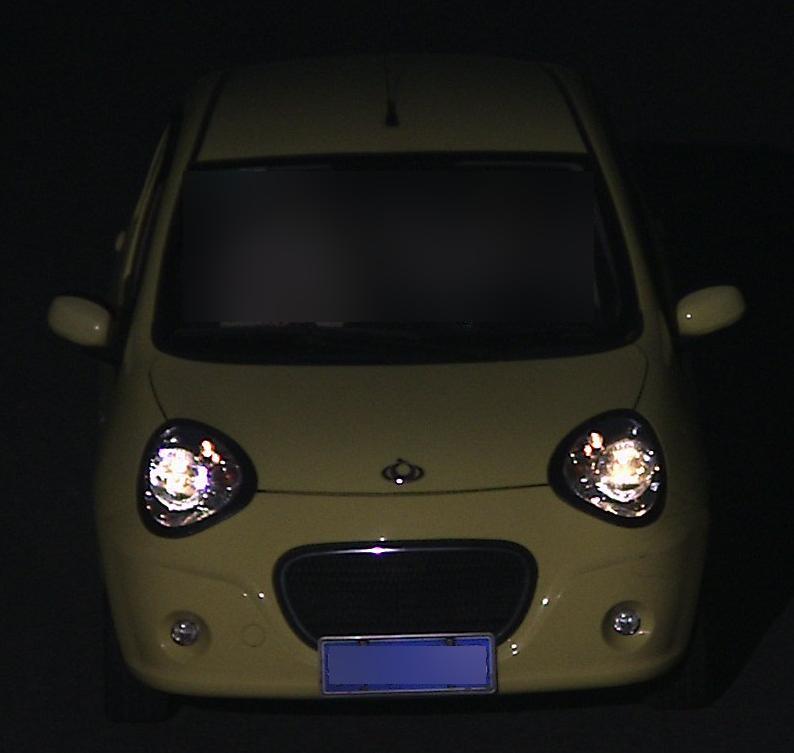}
                \includegraphics[height=12.75ex]{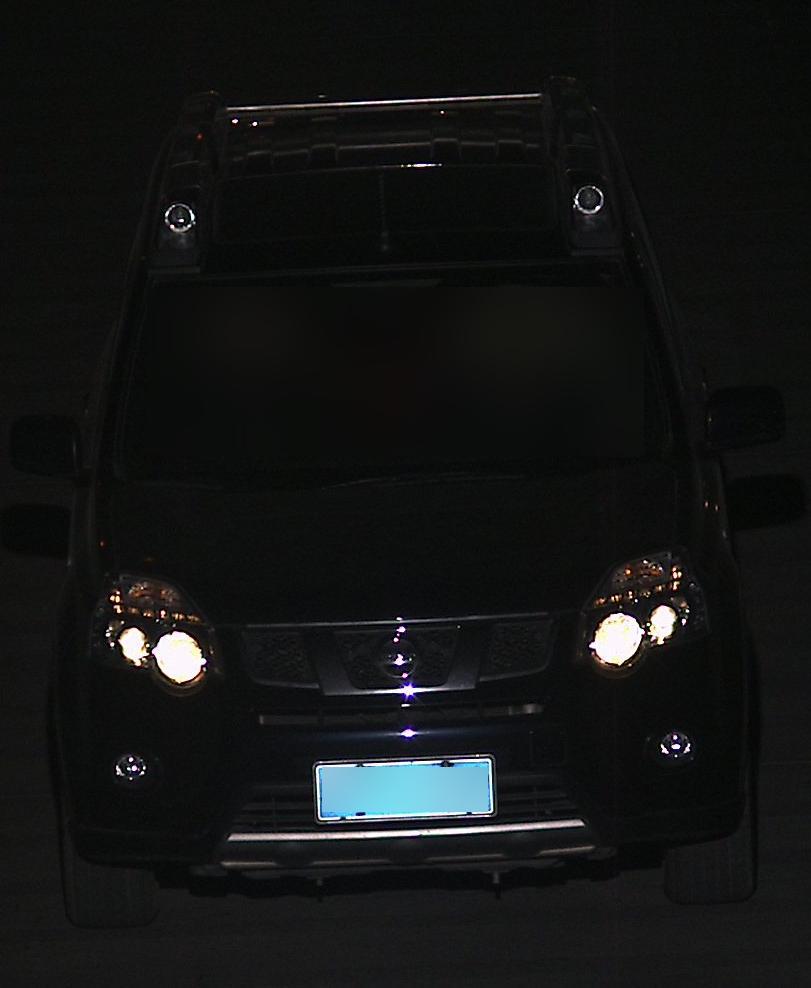}
        }
    }
    
    \vspace{0.25cm} 
    
    \resizebox{0.9\linewidth}{!}{
        \subfloat[]{
                \centering
                
                \includegraphics[height=9ex]{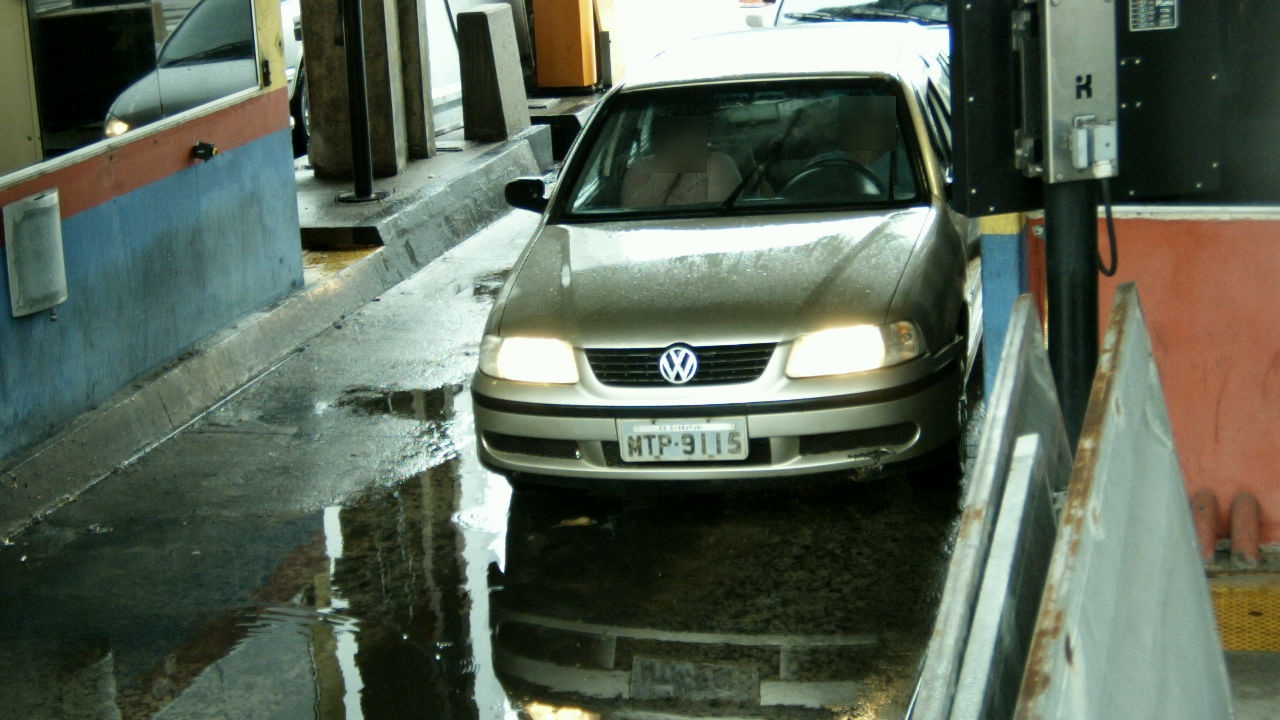}
                \includegraphics[height=9ex]{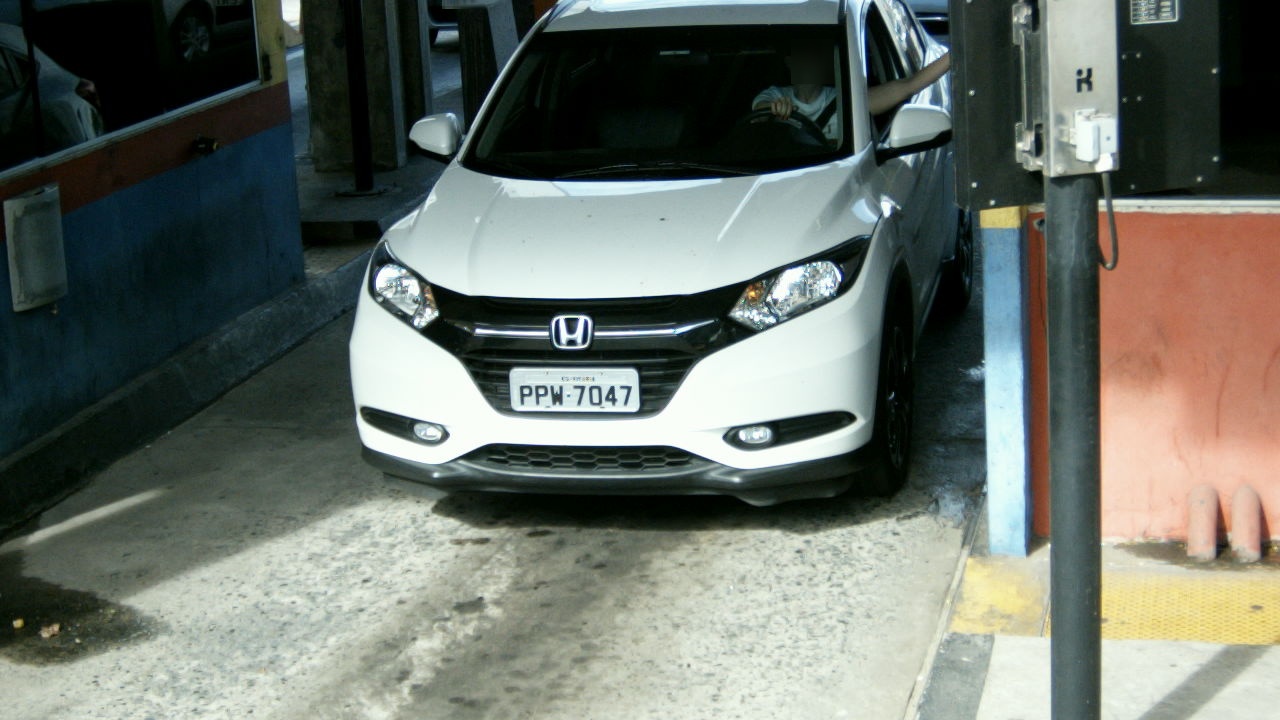}
                \includegraphics[height=9ex]{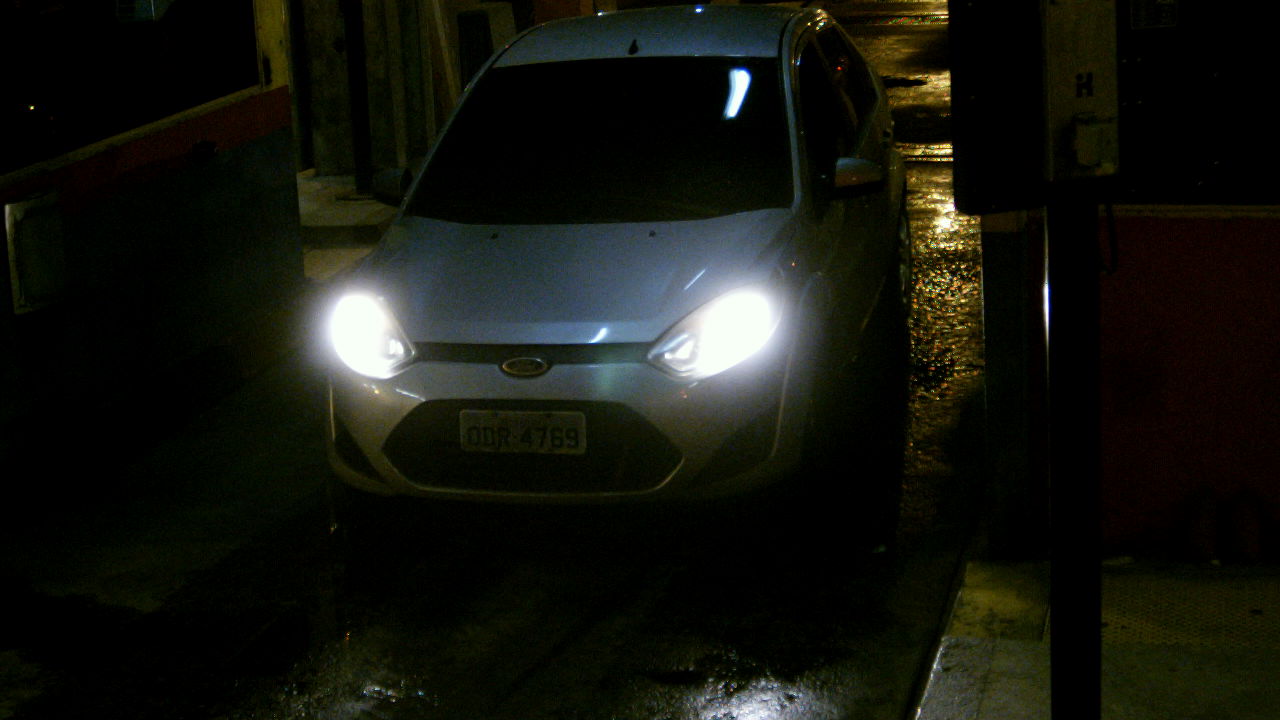}
        }
    }
    
    \vspace{-3.75mm}

    \resizebox{0.9\linewidth}{!}{
        \subfloat[(c)~Images from RodoSol-ALPR, proposed by \cite{laroca2022cross}.]{
                \includegraphics[height=9ex]{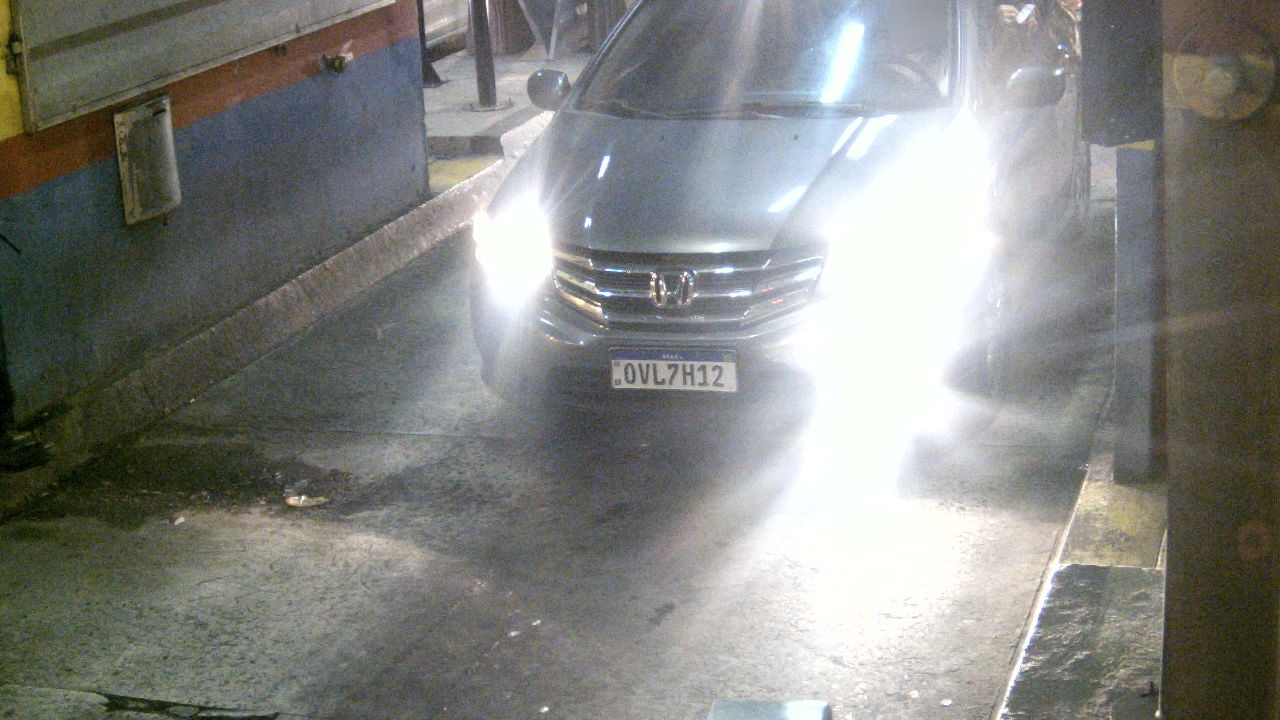}
                \includegraphics[height=9ex]{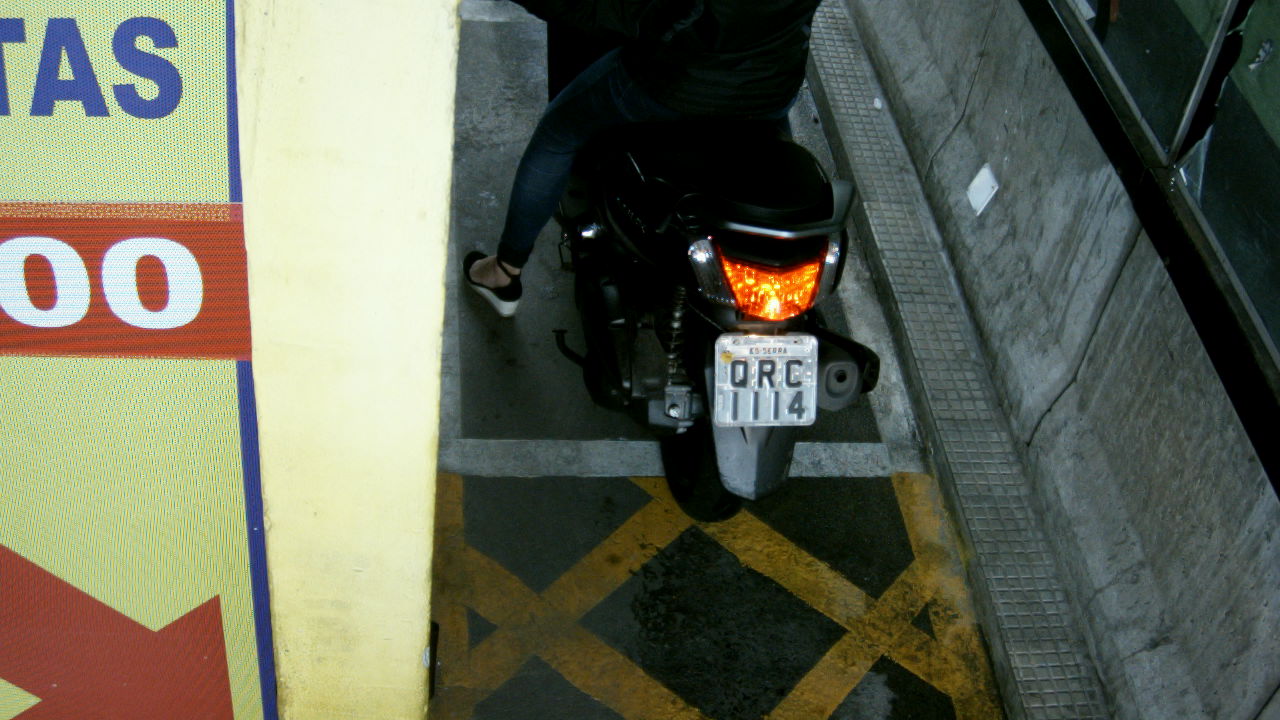}
                \includegraphics[height=9ex]{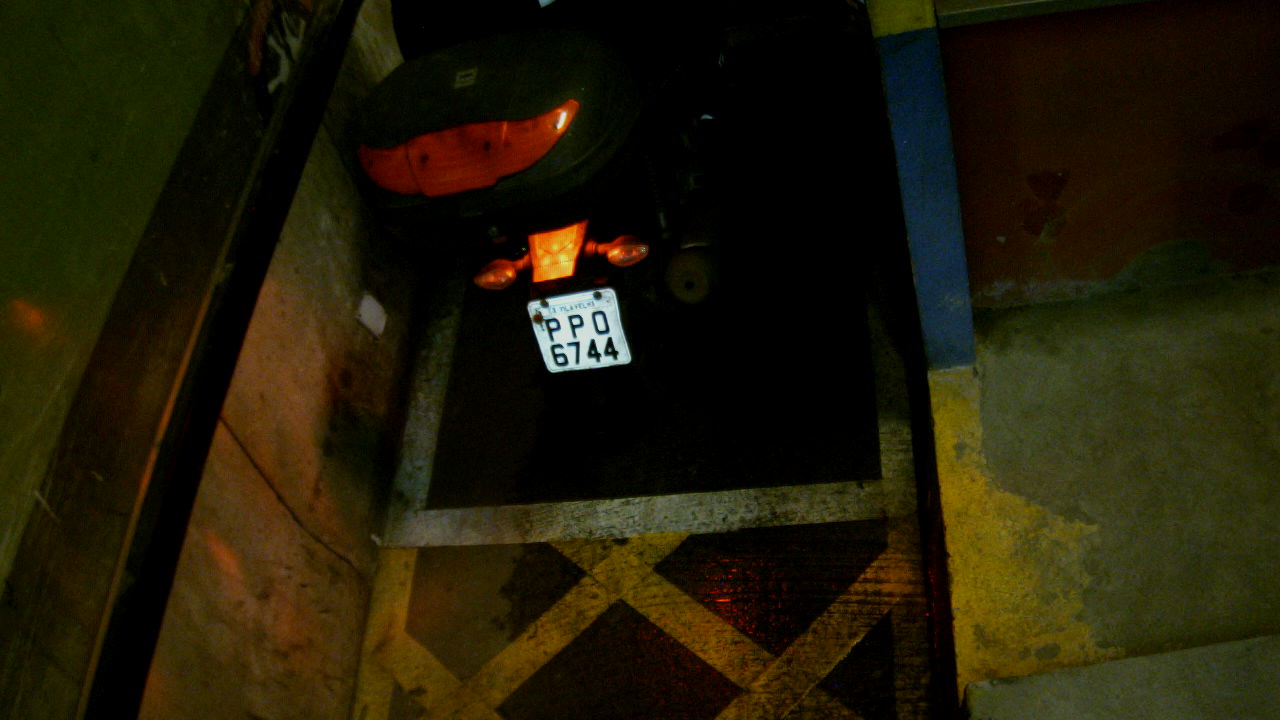}
        }
    }
    
    \vspace{0.25cm} 
    
    \resizebox{0.9\linewidth}{!}{
        \subfloat[]{              
                \includegraphics[height=11.75ex]{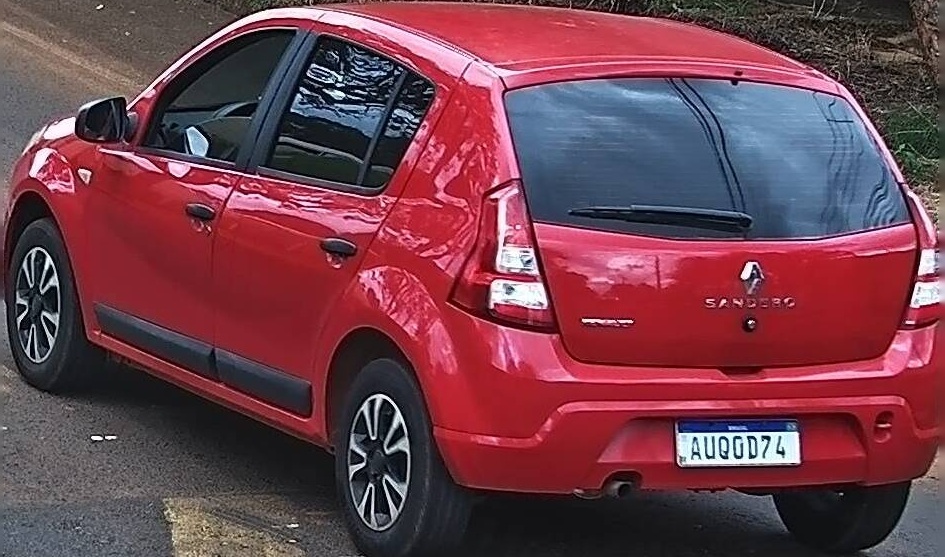}
                \includegraphics[height=11.75ex]{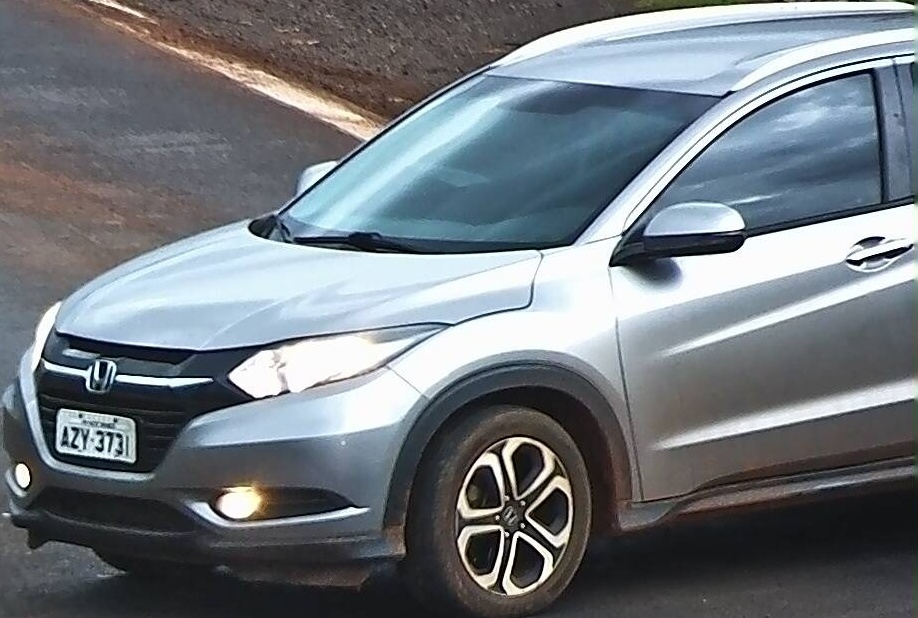}
                \includegraphics[height=11.75ex]{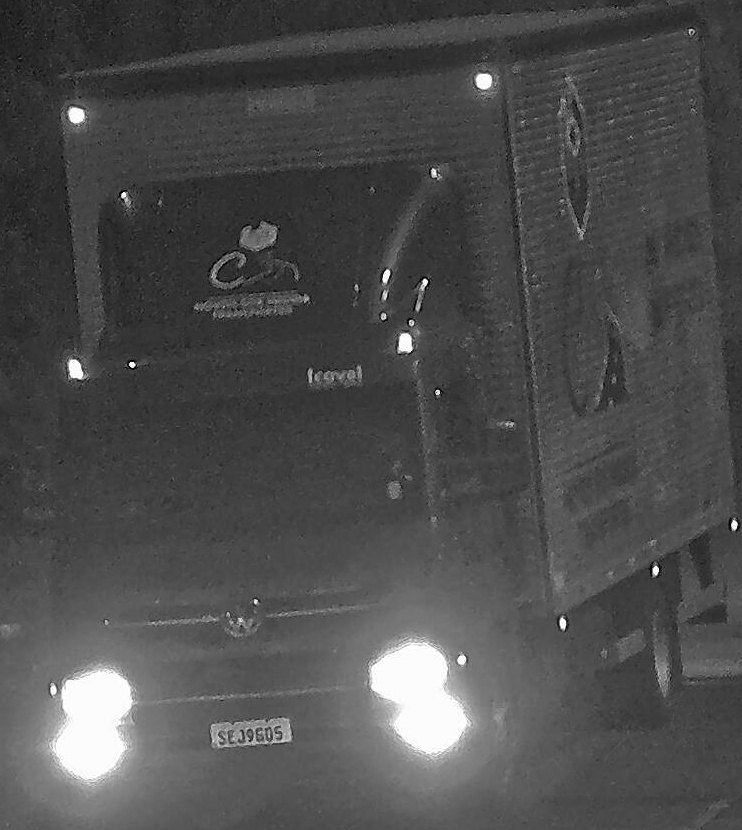}
        }
    }

    \vspace{-3.75mm}

    \resizebox{0.9\linewidth}{!}{
        \subfloat[]{
                \includegraphics[height=11.75ex]{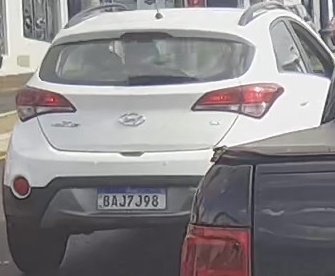}
                \includegraphics[height=11.75ex]{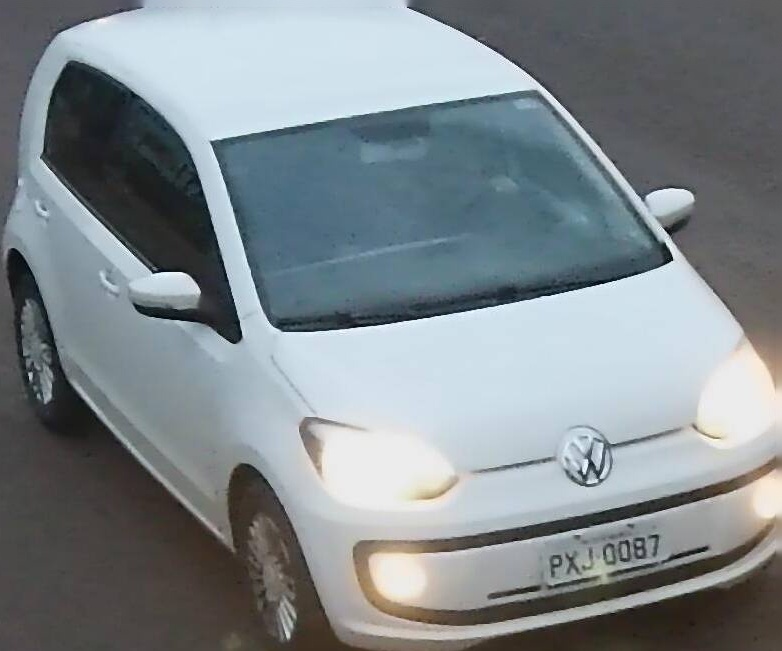}
                \includegraphics[height=11.75ex]{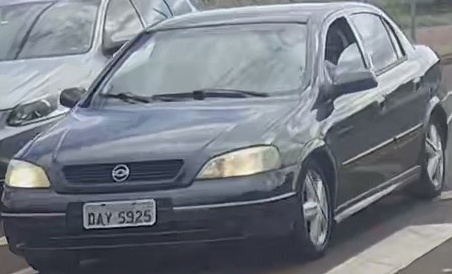}
        }
    }

    \vspace{-3.75mm}

    \resizebox{0.9\linewidth}{!}{
        \subfloat[(d)~Images from \dataset, proposed in this work.]{
                \includegraphics[height=11.75ex]{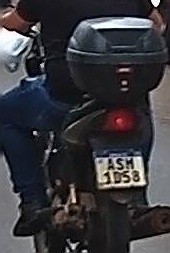}
                \includegraphics[height=11.75ex]{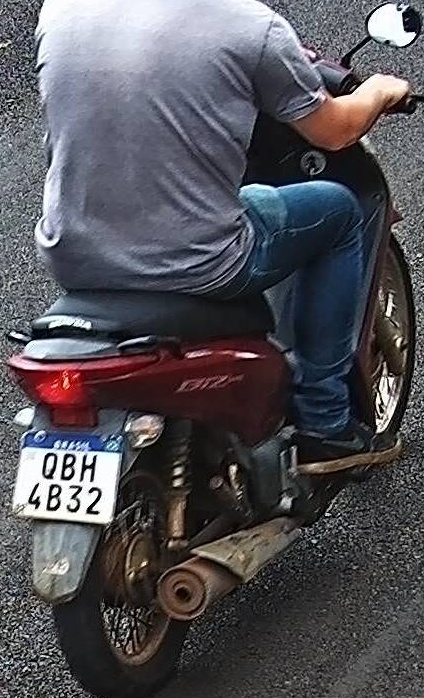}
                \includegraphics[height=11.75ex]{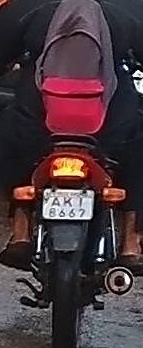}
                \includegraphics[height=11.75ex]{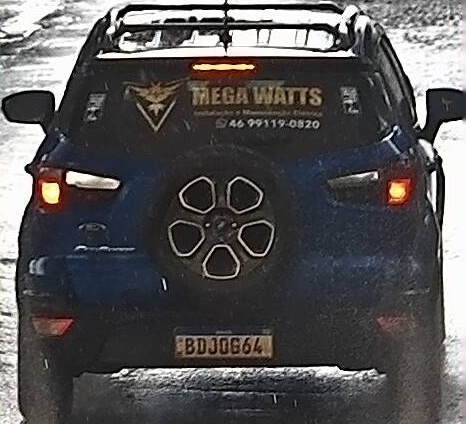}
                \includegraphics[height=11.75ex]{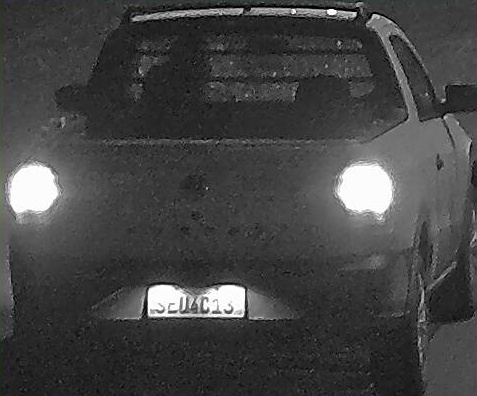}
                
        }
    }
    
    \vspace{0.5mm}

    \caption{Representative images from three public datasets and \dataset. Our dataset features significantly more challenging scenarios, with vehicles captured from diverse viewpoints, environments, lighting conditions, image quality levels, and nighttime infrared~imaging.}
    \label{fig:rw_dataset_comparison}
\end{figure}

\REWRITTEN{
A common limitation among existing datasets is the lack of scenario diversity (illustrated in \cref{fig:rw_dataset_comparison}).
Datasets for vehicle color recognition are collected under controlled conditions, creating optimistic conditions that artificially boost recognition accuracy~\citep{lima2024toward}.
Similarly, while other established datasets (e.g., CompCars-SV and RodoSol-ALPR) incorporate variations in weather and time, they still lack diversity in viewpoints and capture locations.
In contrast, \dataset introduces more diverse and challenging scenarios, capturing vehicles under varying illumination, partial occlusions, and complex real-world conditions. 
}

\subsection{Quantitative Analysis}
\label{subsec:quantitative_analysis}

This section presents an empirical evaluation designed to highlight the limited diversity and lack of challenging scenarios in related datasets. 
To this end, we benchmarked five deep learning classifiers on two widely used datasets -- \cite{chen2014vehicle} dataset for color recognition and CompCars-SV~\citep{yang2015compcars} for model recognition -- and contrasted their performance against the \dataset dataset.

The evaluation considered five architectures: EfficientNet-V2~\citep{tan2021efficientnetv2}, MobileNet-V3~\citep{howard2019mobilenetv3}, ResNet-50~\citep{he2016residual}, Swin Transformer-V2~\citep{liu2022swin}, and \gls*{vit}-b16~\citep{dosovitskiy2021vit}. 
They were chosen due to their strong performance in computer vision tasks, broad support across deep learning frameworks, and adoption in related work~\citep{hassan2021empirical,kuhn2021brcars,lima2024toward}.

To isolate the effect of dataset complexity, a controlled transfer-learning strategy was employed, inspired by our prior work~\citep{lima2024toward}. 
This approach establishes a baseline for comparison. 
Each classifier was initialized with ImageNet-pretrained weights, and only the final classification layer was trained to adapt to the target classes.

All classifiers were trained for up to $500$ epochs using the Adam optimizer ($\beta_1 = 0.9$, $\beta_2 = 0.999$), Cross-entropy loss, a batch size of $128$, and a weight decay of $10^{-5}$. 
We used an initial learning rate of $10^{-3}$, reduced by a factor of $10$ after ten epochs of stagnant validation performance; an early-stopping condition was triggered after $15$ epochs without improvement. 
All images were resized to a $224\times224$ input, preserving the aspect ratio by scaling the longest side and padding the shorter side. 
The training set was subjected to an additional data augmentation pipeline (random rotation, scaling, shearing, brightness/contrast adjustments, motion blur, and random masking\footnote{A file specifying all parameters of the data augmentation pipeline will be included with the dataset release.}).
As the final step for all images, normalization was applied using the standard ImageNet mean and standard deviation.

The evaluation was conducted using specific protocols for each dataset: our custom splitting protocol for \dataset (see~\cref{subsec:splitting_protocol}), the methodology from our prior work~\citep{lima2024toward} for \cite{chen2014vehicle} dataset, and the original protocol for CompCars~\citep{yang2015compcars}. 
Classification performance was measured using macro accuracy (Ma-Acc), micro accuracy (Mi-Acc), and F1-score (F1).
Micro accuracy reflects the overall performance, while macro accuracy provides a balanced measure that gives equal weight to both frequent and rare classes.

The results in \cref{tab:qualitative_analysis_results} reveal a clear performance gap. 
The simple transfer-learning approach was effective on the reference benchmarks; ViT-b16 achieved over $92$\% micro-accuracy, a result close to literature reports~\citep{hu2023vehicle,amirkhani2023deepcar}. 
In contrast, classifiers performed worse on \dataset dataset due to its challenging nature and more realistic evaluation setting.
This highlights that benchmarks with limited diversity risk producing optimistic evaluations that do not reflect practical performance. 

\begin{table}
    \caption{Performance metrics (\%) achieved by all classifiers using the transfer-learning approach on the \cite{chen2014vehicle} dataset (vehicle color recognition), the CompCars dataset~\citep{yang2015compcars} (vehicle model recognition), and the proposed \dataset dataset (vehicle color, make, model, and type recognition). Results on the proposed dataset are averaged over ten runs, with standard deviations shown in parentheses.}
    \label{tab:qualitative_analysis_results}
    \resizebox{0.998\linewidth}{!}{
        \begin{tabular}{lcccccc}
            \multicolumn{7}{l}{Results for color~\citep{chen2014vehicle} and model~\citep{yang2015compcars} recognition benchmarks.} \\
            \toprule
            \multicolumn{1}{l}{\multirow{2}{*}{Classifier}} &
              \multicolumn{3}{c}{\textit{\cite{chen2014vehicle}}} &
              \multicolumn{3}{c}{\textit{CompCars~\citep{yang2015compcars}}} \\
            \multicolumn{1}{c}{} &
              Mi-Acc &
              Ma-Acc &
              F1 &
              Mi-Acc &
              Ma-Acc &
              F1 \\
            \midrule
            EfficientNet-V2 &
              $85.5$ &
              $85.6$ &
              $86.8$ &
              $73.5$ &
              $70.5$ &
              $72.2$ \\
            MobileNet-V3 &
              $88.1$ &
              $90.2$ &
              $90.4$ &
              $82.0$ &
              $79.2$ &
              $81.2$ \\
            ResNet-50 &
              $92.7$ &
              $93.4$ &
              $93.8$ &
              $87.4$ &
              $82.5$ &
              $85.0$ \\
            Swin Transformer-V2 &
              $91.8$ &
              $93.0$ &
              $93.6$ &
              $92.4$ &
              $89.0$ &
              $90.1$ \\
            ViT-b16 &
              $\mathbf{92.9}$ &
              $\mathbf{93.5}$ &
              $\mathbf{94.0}$ &
              $\mathbf{94.6}$ &
              $\mathbf{92.9}$ &
              $\mathbf{93.9}$ \\
            \bottomrule
            & & & & & & \\ 
            \\
            \multicolumn{7}{l}{Results on the UFPR-VeSV dataset.} \\
            \toprule
            \multicolumn{1}{l}{\multirow{2}{*}{Classifier}} &
              \multicolumn{3}{c}{\textit{UFPR-VeSV Color}} &
              \multicolumn{3}{c}{\textit{UFPR-VeSV Type}} \\
            \multicolumn{1}{c}{} &
              Mi-Acc &
              Ma-Acc &
              F1 &
              Mi-Acc &
              Ma-Acc &
              F1 \\
            \midrule
            EfficientNet-V2 &
              $77.8$ $(0.6)$ &
              $46.1$ $(1.7)$ &
              $47.0$ $(1.8)$ &
              $75.2$ $(0.5)$ &
              $55.6$ $(2.7)$ &
              $59.7$ $(2.7)$ \\
            MobileNet-V3 &
              $81.0$ $(0.5)$ &
              $50.9$ $(1.4)$ &
              $53.2$ $(1.6)$ &
              $76.9$ $(0.3)$ &
              $56.4$ $(2.0)$ &
              $61.4$ $(2.6)$ \\
            ResNet-50 &
              $83.2$ $(0.5)$ &
              $54.2$ $(0.8)$ &
              $56.6$ $(1.1)$ &
              $81.7$ $(0.4)$ &
              $65.4$ $(1.7)$ &
              $70.4$ $(1.4)$ \\
            Swin Transformer-V2 &
              $\mathbf{87.2}$ $\mathbf{(0.4)}$ &
              $56.5$ $(1.5)$ &
              $60.2$ $(1.7)$ &
              $84.4$ $(0.7)$ &
              $\mathbf{72.8 (1.8)}$ &
              $\mathbf{76.8 (1.9)}$ \\
            ViT-b16 &
              $\mathbf{87.5}$ $\mathbf{(0.4)}$ &
              $\mathbf{60.2}$ $\mathbf{(1.8)}$ &
              $\mathbf{63.3}$ $\mathbf{(1.6)}$ &
              $\mathbf{86.3}$ $\mathbf{(0.7)}$ &
              $\mathbf{75.0}$ $\mathbf{(2.7)}$ &
              $\mathbf{77.1}$ $\mathbf{(2.3)}$ \\
            \bottomrule
            \\
            \toprule
            \multicolumn{1}{l}{\multirow{2}{*}{Classifier}} &
              \multicolumn{3}{c}{\textit{UFPR-VeSV Make}} &
              \multicolumn{3}{c}{\textit{UFPR-VeSV Model}} \\
            \multicolumn{1}{c}{} &
              Mi-Acc &
              Ma-Acc &
              F1 &
              Mi-Acc &
              Ma-Acc &
              F1 \\
            \midrule
            EfficientNet-V2 &
              $42.6$ $(0.6)$ &
              $23.7$ $(0.9)$ &
              $24.1$ $(0.9)$ &
              $35.3$ $(0.7)$ &
              $21.5$ $(0.8)$ &
              $22.6$ $(0.8)$ \\
            MobileNet-V3 &
              $45.7$ $(1.0)$ &
              $26.3$ $(1.3)$ &
              $28.4$ $(1.4)$ &
              $40.8$ $(0.7)$ &
              $25.2$ $(1.0)$ &
              $27.7$ $(1.2)$ \\
            ResNet-50 &
              $51.0$ $(1.3)$ &
              $30.9$ $(1.3)$ &
              $33.4$ $(1.3)$ &
              $46.8$ $(0.9)$ &
              $28.1$ $(0.7)$ &
              $30.4$ $(0.9)$ \\
            Swin Transformer-V2 &
              $51.6$ $(1.0)$ &
              $32.0$ $(1.2)$ &
              $32.7$ $(1.4)$ &
              $49.8$ $(0.7)$ &
              $32.8$ $(1.1)$ &
              $34.8$ $(1.0)$ \\
            ViT-b16 &
              $\mathbf{58.1}$ $\mathbf{(0.5)}$ &
              $\mathbf{39.1}$ $\mathbf{(1.1)}$ &
              $\mathbf{42.0}$ $\mathbf{(1.0)}$ &
              $\mathbf{57.4}$ $\mathbf{(1.3)}$ &
              $\mathbf{40.8}$ $\mathbf{(1.1)}$ &
              $\mathbf{43.7}$ $\mathbf{(1.4)}$ \\
            \bottomrule
        \end{tabular}
    }
\vspace{0.5mm}
\end{table}

%% file: 5-fine-grained.tex
\section{Fine-grained Vehicle Classification}
\label{sec:FGVC}

This section establishes the \gls*{fgvc} performance baseline on the \dataset dataset. 
The analysis is split into two parts. 
In \cref{subsec:fgvc_singletask}, we benchmark five deep learning classifiers for color, make, model, and type recognition and analyze the results for each task individually. 
In \cref{subsec:fgvc_joint}, we use the predictions from the best-performing classifier to analyze the results when these tasks are considered in combination.

\subsection{Single-Task Analysis}
\label{subsec:fgvc_singletask}

This section establishes the \gls*{fgvc} baseline on the \dataset dataset. 
Unlike the transfer-learning strategy in \cref{subsec:quantitative_analysis}, we now employ an end-to-end fine-tuning strategy for five deep learning architectures: EfficientNet-V2, MobileNet-V3, ResNet-50, Swin Transformer-V2, and ViT-b16. 
The splitting protocol, loss function, and evaluation metrics remain consistent with the prior qualitative analysis.

All classifiers were initialized with ImageNet pre-trained weights, and all layers were set as trainable. 
Training was conducted for up to $500$ epochs, with early stopping halting the process after $30$ epochs without validation improvement. 
We used the Adam optimizer ($\beta_1 = 0.9$, $\beta_2 = 0.999$) with a $5\times10^{-4}$ weight decay and a $64$ batch size. 
The initial learning rate was $10^{-2}$, reduced by a factor of $10$ after $20$ epochs of stagnant validation loss.

A comprehensive data augmentation pipeline was applied during training to enhance generalization. 
This included random resized cropping ($224\times224$ pixels), random horizontal flipping, and RandAugment~\citep{cubuk2020randaugment} (two transformations, intensity 9). 
For inference, images were also resized to $224\times224$ while preserving their aspect ratio. 
As a final preprocessing step, all images were normalized using the standard ImageNet mean and standard deviation.

\cref{tab:ufpr_vesv_benchmark} presents the performance of each classifier across the \gls*{fgvc} tasks. 
For the attention-based models (ViT and Swin Transformer), the results shown are from the transfer-learning approach (\cref{subsec:quantitative_analysis}), as this strategy yielded superior performance. 
We attribute this is because the large data requirements of transformers can limit generalization when all layers are fine-tuned.

\begin{table}[!htb]
    \caption{Performance metrics (\%) achieved by all classifiers using the end-to-end fine-tuning approach on vehicle color, make, model, and type recognition (averaged over 10 runs).
Standard deviations are shown in parentheses.}\label{tab:ufpr_vesv_benchmark}
    \resizebox{0.998\linewidth}{!}{
        \begin{tabular}{lcccccc}
            \toprule
            \multicolumn{1}{l}{\multirow{2}{*}{Classifier}} &
              \multicolumn{3}{c}{\textit{UFPR-VeSV Color}} &
              \multicolumn{3}{c}{\textit{UFPR-VeSV Type}} \\
            \multicolumn{1}{c}{} &
              Mi-Acc &
              Ma-Acc &
              F1 &
              Mi-Acc &
              Ma-Acc &
              F1 \\
            \midrule
            EfficientNet-V2 &
              $\mathbf{93.5}$ $\mathbf{(0.6)}$ &
              $\mathbf{71.5}$ $\mathbf{(2.3)}$ &
              $\mathbf{73.8}$ $\mathbf{(2.3)}$ &
              $\mathbf{96.1}$ $\mathbf{(0.7)}$ &
              $\mathbf{89.0}$ $\mathbf{(2.0)}$ &
              $\mathbf{90.2}$ $\mathbf{(1.6)}$ \\
            MobileNet-V3 &
              $\mathbf{93.2}$ $\mathbf{(0.7)}$ &
              $\mathbf{71.2}$ $\mathbf{(2.9)}$ &
              $\mathbf{73.2}$ $\mathbf{(2.5)}$ &
              $95.2$ $(0.7)$ &
              $85.6$ $ (1.9)$ &
              $87.4$ $ (1.6)$ \\
            ResNet-50 &
              $93.1$ $ (0.4)$ &
              $69.6$ $ (2.6)$ &
              $71.8$ $ (2.3)$ &
              $95.5$ $ (0.6)$ &
              $86.8$ $ (2.3)$ &
              $88.4$ $ (1.9)$ \\
            Swin Transformer-V2 &
              $87.2$ $(0.4)$ &
              $56.5$ $(1.5)$ &
              $60.2$ $(1.7)$ &
              $84.4$ $(0.67$ &
              $72.8$ $(1.8)$ &
              $76.8$ $(1.9)$ \\
            ViT-b16 &
              $87.5$ $(0.4)$ &
              $60.2$ $(1.8)$ &
              $63.3$ $(1.6)$ &
              $86.3$ $(0.7)$ &
              $75.0$ $(2.7)$ &
              $77.1$ $(2.3)$ \\
            \bottomrule
            \\
            \toprule
            \multicolumn{1}{l}{\multirow{2}{*}{Classifier}} &
              \multicolumn{3}{c}{\textit{UFPR-VeSV Make}} &
              \multicolumn{3}{c}{\textit{UFPR-VeSV Model}} \\
            \multicolumn{1}{c}{} &
              Mi-Acc &
              Ma-Acc &
              F1 &
              Mi-Acc &
              Ma-Acc &
              F1 \\
            \midrule
            EfficientNet-V2 &
              $\mathbf{94.4}$ $\mathbf{(0.6)}$ &
              $\mathbf{85.0}$ $\mathbf{(1.6)}$ &
              $\mathbf{86.4}$ $\mathbf{(1.4)}$ &
              $\mathbf{90.9}$ $\mathbf{(0.6)}$ &
              $\mathbf{86.2}$ $\mathbf{(1.1)}$ &
              $\mathbf{87.3}$ $\mathbf{(0.8)}$ \\
            MobileNet-V3 &
              $91.3$ $(0.7)$ &
              $78.0 $ $1.7)$ &
              $80.2 $ $1.4)$ &
              $86.5$ $(0.8)$ &
              $79.2$ $(1.2)$ &
              $80.9$ $(1.1)$ \\
            ResNet-50 &
              $93.6$ $(0.5)$ &
              $83.5$ $(1.7)$ &
              $85.0$ $(1.1)$ &
              $89.9$ $(0.9)$ &
              $84.6$ $(1.2)$ &
              $85.7$ $(0.8)$ \\
            Swin Transformer-V2 &
              $51.6$ $(1.0)$ &
              $32.0$ $(1.2)$ &
              $32.7$ $(1.4)$ &
              $49.8$ $(0.7)$ &
              $32.8$ $(1.1)$ &
              $34.8$ $(1.0)$ \\
            ViT-b16 &
              $58.1$ $(0.5)$ &
              $39.1$ $(1.1)$ &
              $42.0$ $(1.0)$ &
              $57.4$ $(1.3)$ &
              $40.8$ $(1.1)$ &
              $43.7$ $(1.4)$ \\
            \bottomrule
        \end{tabular}
    }
\end{table}

A general trend across all classifiers was the significant gap between micro-accuracy and the macro metrics (macro-accuracy and F1-score). 
This discrepancy is an expected result of the dataset’s severe class imbalance, which causes classifiers to perform well on frequent classes but struggle with underrepresented ones.
With this in mind, we selected EfficientNet-V2 for a detailed analysis to identify the specific challenges for each task, as it achieved the best performance across the experiments. 

\begin{figure}[!htb]
    \centering
    \captionsetup[subfigure]{captionskip=1.5pt,font=footnotesize}
    \subfloat[Red]{\includegraphics[height=15ex]{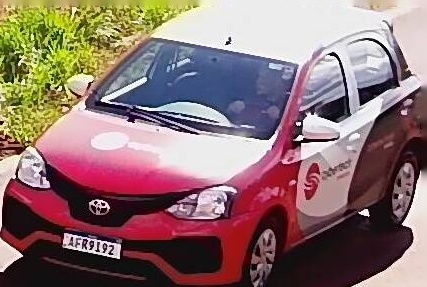}}
    \hspace{0.2mm}
    \subfloat[Black]{\includegraphics[height=15ex]{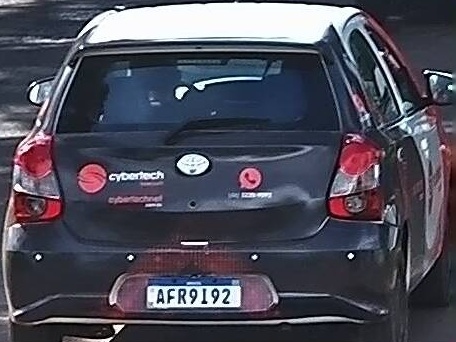}}
    \caption{Examples of misclassified multicolored vehicle from different viewpoints. The predicted color is shown below each image. Depending on the camera angle and illumination, a single color can appear dominant, which causes the classifier to classify the vehicle based on that one color.}
    \label{fig:error_multicolored}
\end{figure}

In color recognition, the classifier performed worst for beige, brown, blue, green, and gray. 
This is likely due to lighting variations and close shade similarities, such as darker shades appearing black or lighter beige appearing silver. 
The ``multicolored'' class was also challenging, with the classifier often predicting one of the vehicle's colors. 
This error is attributed to image perspective and illumination conditions that can make one color appear dominant (see~\cref{fig:error_multicolored}).

For make recognition, the ``others'' class had the lowest performance ($\approx30$\%). 
This suggests that a superclass for less common makes is ineffective, highlighting the need for out-of-distribution methods.
Another common source of error was inter-manufacturer confusion for similar vehicle types; 
for example, Land-Rover which primarily sells SUVs in Brazil was often misclassified by more representative SUV manufacturers. 

\begin{figure}[!htb]
    \centering
    \captionsetup[subfigure]{captionskip=1.5pt,font=footnotesize}
    \subfloat[Peugeot Boxer]{\includegraphics[height=17ex]{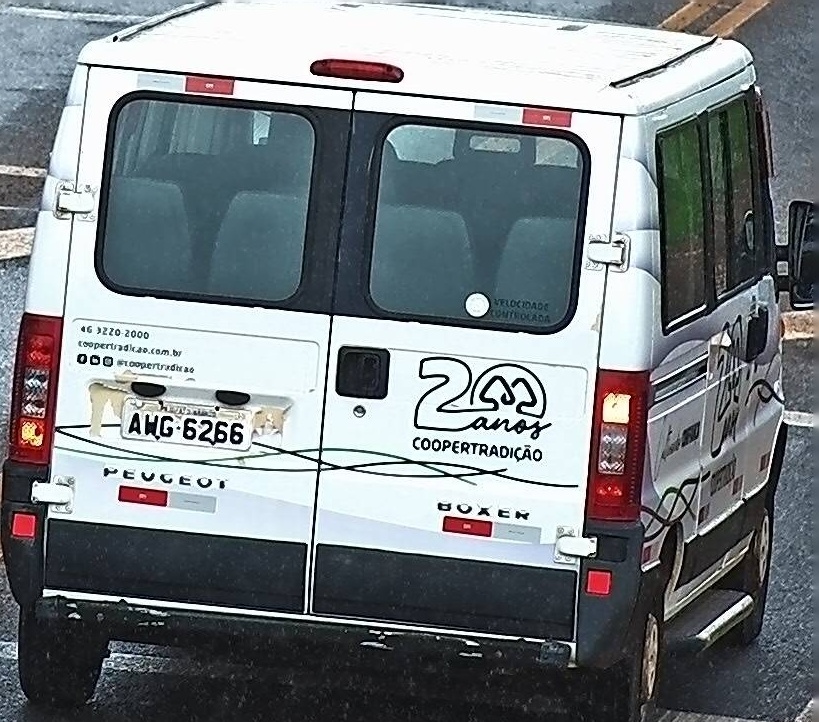}}
    \hspace{0.2mm}
    \subfloat[Fiat Ducato]{\includegraphics[height=17ex]{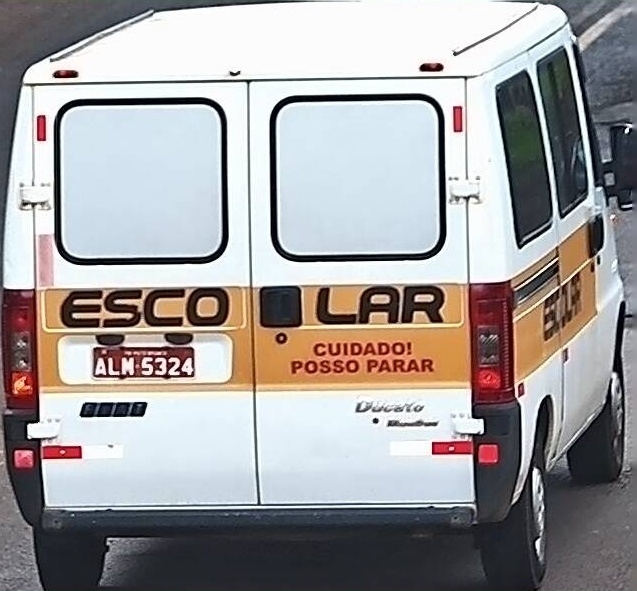}}
    \caption{Examples of misclassified vehicle models from different manufacturers. Original make and model are displayed below each image. In~(a), a rear-view Boxer was misclassified as a Ducato, while in~(b) a rear-view Ducato was misclassified as a Boxer. Aside from minor brand-specific markings, the vehicles have a very similar structure, making accurate differentiation~challenging.}
    \label{fig:error_similar_models}
\end{figure}

In vehicle model recognition, three primary sources of error were identified. 
First, a single vehicle platform is often sold in multiple body-style variants (e.g., sedan, hatch, compact pickup) that are visually similar, especially from the front.
Second, some models are sold under different names by different manufacturers in Brazil (see~\cref{fig:error_similar_models}). 
Finally, a consistent design language across different models from the same make also contributed to errors.

In type recognition, misclassifications were frequent between related types, such as scooter versus motorcycle and truck classes (i.e., compact-truck, tractor-truck, and truck). 
Another common error was the confusion of semi-trailers with trucks, as their images typically include the towing truck.
Finally, compact-pickups and cars were also confused when derived from the same base vehicle platform.

The preceding analysis highlighted the difficulty of distinguishing similar vehicles under challenging real-world \gls*{fgvc} conditions. 
However, relying on a single prediction is often insufficient for practical applications, especially in public security (e.g., criminal investigation or forensics), where a limited set of high-probability candidates is more useful than a single classification. 
Therefore, the performance of EfficientNet-V2 was further analyzed using top-1, top-2, and top-3 predictions, as detailed in \cref{tab:fgvc_topk}.

\begin{table}[!htp]\centering
    \caption{Top-1, top-2, and top-3 performance metrics (\%) for EfficientNet-V2 on vehicle color, make, model, and type recognition (\dataset). Results are reported as the mean over 10 runs. Standard deviations are in parentheses.}
    \label{tab:fgvc_topk}
    \resizebox{0.99\linewidth}{!}{
        \begin{tabular}{lccccccccc}
            \toprule
            \multirow{2}{*}{Task} &
            \multicolumn{3}{c}{Top-1} &
            \multicolumn{3}{c}{Top-2} &
            \multicolumn{3}{c}{Top-3} \\
            \cmidrule{2-10}
            &Mi-Acc &Ma-Acc &F1 &Mi-Acc &Ma-Acc &F1 &Mi-Acc &Ma-Acc &F1 \\
            \midrule
            Color &$93.5$ $(0.6)$ &$71.5$ $(2.3)$ &$73.8$ $(2.3)$ &$98.0$ $(0.3)$ &$87.0$ $(1.7)$ &$89.4$ $(1.5$) &$99.1$ $(0.1)$ &$93.5$ $(1.4)$ &$95.0$ $(1.0)$ \\
            Make &$94.4$ $(0.6)$ &$85.0$ $(1.6)$ &$86.4$ $(1.4)$ &$96.8$ $(0.4)$ &$90.8$ $(1.3)$ &$91.6$ $(1.3)$ &$97.8$ $(0.3)$ &$92.9$ $(1.3)$ &$93.7$ $(1.2)$ \\
            Model &$90.9$ $(0.6)$ &$86.2$ $(1.1)$ &$87.3$ $(0.8)$ &$95.3$ $(0.5)$ &$91.5$ $(1.1)$ &$92.3$ $(0.9)$ &$96.7$ $(0.5)$ &$93.8$ $(0.9)$ &$94.5$ $(0.7)$ \\
            Type &$96.1$ $(0.7)$ &$89.0$ $(2.0)$ &$90.2$ $(1.6)$ &$99.2$ $(0.2)$ &$97.2$ $(1.2)$ &$97.7$ $(0.8)$ &$99.7$ $(0.1)$ &$98.9$ $(0.8)$ &$99.0$ $(0.6)$ \\
            \bottomrule
        \end{tabular}
    }
\end{table}

The improvement in top-k metrics demonstrates that the correct classification is frequently included within the top three candidates. 
This suggests the classifier is partially robust to the dataset's inherent ambiguities, a finding with practical implications for the deployment of real-world systems. 
However, substantial room for improvement remains, particularly in the macro-accuracy metrics for color, make, and model recognition.

A broader analysis of classification errors revealed that nighttime infrared images were a principal source of misclassification. 
Despite representing only $21.5$\% of the dataset, these images contributed to $53.4$\%, $42.9$\%, and $39.4$\% of total top-1 errors for make, model, and type recognition, respectively.
The issue persisted even in the top-3 analysis, where this condition accounted for $\approx60$\% of the remaining errors for the same tasks. 

In contrast to the clear negative impact of infrared conditions, viewpoint had a mixed effect on performance.
The frontal view improved make recognition, likely due to the visibility of the manufacturer's badge. 
Conversely, this view was less effective for model and type recognition, as it offers fewer distinguishing features.

\subsection{Joint-Task Analysis}
\label{subsec:fgvc_joint}

The previous section established baselines by evaluating the color, make, model, and type recognition tasks in isolation. 
However, this information is semantically connected: a hierarchical relationship exists between make, model, and type, while color is an orthogonal property. 
Moreover, in practical surveillance applications, queries can range from a single attribute (e.g., ``blue cars'') to a complex, multi-attribute conjunction (e.g., ``blue Ford models''). 

Therefore, this work also evaluates simultaneous accuracy to measure performance on these conjunctive tasks. 
Using the predictions from the previously trained EfficientNet-V2 classifiers~(the best performing method), this metric is defined as the percentage of images for which all attributes within a specified set are correctly predicted. 
\cref{tab:fgvc_hierarchy} illustrates how this metric changes as the set of required attributes expands. 
The expansion follows the logical ``type-make-model'' hierarchy, with the independent ``color'' attribute included as the final component.

\begin{table}[!htp]\centering
    \caption{EfficientNet-V2 simultaneous accuracy (\%) for conjunctive attribute recognition, detailing the performance as the set of required attributes expands. Results are reported as mean (standard deviation in parentheses) over 10 runs.}
    \label{tab:fgvc_hierarchy}
    \resizebox{0.65\linewidth}{!}{
        \begin{tabular}{lc}
        \toprule
        Attribute set &Accuracy \\
        \midrule
        Type                            &$96.1$ $(0.7)$ \\
        Type \& Make                    &$91.5$ $(1.0)$ \\
        Type \& Make \& Model           &$85.5$ $(1.1)$ \\
        Type \& Make \& Model \& Color  &$80.2$ $(1.2)$ \\
        \bottomrule
        \end{tabular}
    }
\end{table}

As shown in \cref{tab:fgvc_hierarchy}, simultaneous accuracy predictably degrades as more attributes are required, a drop attributed to the compounding of individual error rates. 
This highlights a significant gap: while single-task accuracy can exceed $90$\% (\cref{tab:ufpr_vesv_benchmark}, \cref{subsec:fgvc_singletask}), the ability to produce a simultaneously correct vehicle description remains a challenge, underscoring the need for improved joint-task recognition.

Beyond this performance degradation, the isolated training/evaluation approach introduces a more critical issue: logically inconsistent predictions. 
The vehicle attributes are semantically bound; for example, the ``Fiat Ducato'' model inherently belongs to the ``Fiat'' make. 
Because the classifiers were trained independently, they are unaware of these dependencies. 
This leads to nonsensical outputs, such as $2.9$\% of predictions where the model was correctly identified, but its corresponding make was~not.

\begin{figure}[!htb]
    \centering
    \captionsetup[subfigure]{captionskip=1.5pt,font=footnotesize}
    \subfloat[Original]{\includegraphics[height=14ex]{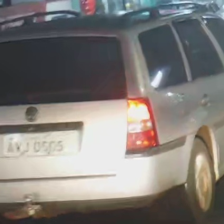}}
    \hspace{0.2mm}
    \subfloat[Make]{\includegraphics[height=14ex]{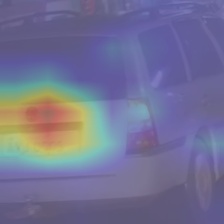}}
    \hspace{0.2mm}
    \subfloat[Model]{\includegraphics[height=14ex]{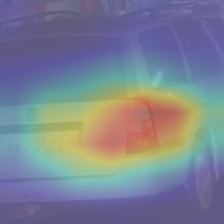}}
    \caption{Grad-CAM~\citep{selvaraju2017gradcam} attention maps for a Volkswagen Parati. (a) Original image. (b) The make recognition map, focusing on the manufacturer’s badge. (c) The model recognition map, relying on other features, such as the headlights.}
    \label{fig:gradcam}
\end{figure}

Such inconsistencies represent a critical failure in a practical system. 
This issue is explained by the classifiers learning different, independent features for each task. 
We confirm this divergence by Grad-CAM~\citep{selvaraju2017gradcam} attention map analysis, which shows that the make recognition focuses primarily on the manufacturer’s badge, while the model recognition relies on other features, such as headlights (see an example in \cref{fig:gradcam}).
Thus, make recognition could fail if the badge is hidden or blurry, while model recognition can still succeed if its required features remain visible.

This analysis reveals a key limitation for practical \gls*{fgvc}: isolated models are insufficient, as they can produce logically inconsistent results. 
Methods must, therefore, enforce hierarchical consistency.
Multi-task learning is a promising solution, as it would require the classifier to learn shared representations that explicitly encompass this hierarchy, which would reduce or eliminate such contradictions.

%% file: 6-alpr.tex
\section{ALPR and FGVC Integration}
\label{sec:ALPR}

This section evaluates \gls*{alpr} and its integration with \gls*{fgvc}. 
These two tasks are crucial for vehicle identification but are often studied independently, missing their combined potential. 
First, \cref{subsec:lpr} establishes an \gls{lpr} baseline by assessing the performance of two state-of-the-art optical character recognition methods on the \dataset dataset. 
Following this, \cref{subsec:lpr_fgvc_integration} analyzes the potential of a joint system, quantifying how \gls*{fgvc} can serve as a support mechanism to enhance \gls*{alpr} robustness.

\subsection{License Plate Recognition}
\label{subsec:lpr}

This experiment assessed the performance of two optical character recognition methods on the \glspl*{lp} extracted from the \dataset images: GP-\gls*{alpr}~\citep{liu2024irregular} and ParSeq-Tiny~\citep{bautista2022str}. 
GP-\gls*{alpr} is specifically designed to handle irregular \glspl*{lp} using deformable spatial attention and global perception modules. 
ParSeq-Tiny is a general scene text recognition method that combines context-free non-autoregressive and context-aware autoregressive inference through permutation language~modeling.

These models were selected  for their state-of-the-art performance, widespread adoption in prior research~\citep{nascimento2024enhancing,du2025instruction}, and public availability, which facilitates reproducibility.
Both models were trained from scratch in accordance with the original methodologies proposed by their respective authors. 
Their performance was evaluated using two metrics: \gls*{lp}-level accuracy, which reflects the percentage of \glspl*{lp} correctly recognized in their entirety, and character-level accuracy, which measures the proportion of individual characters accurately identified.

\cref{tab:lpr_results} compares the \gls*{lpr} results for GP-\gls*{alpr} and ParSeq-Tiny. 
The latter achieved the highest performance, with an average \gls*{lp}-level accuracy of \major{$98.0$\%}. 
This high accuracy was an expected outcome;
\gls*{fgvc} annotations were primarily based on official vehicle data retrieved using the \glspl*{lp}, thus ensuring that all \glspl*{lp} are human-recognizable in some form.

\begin{table}[!htp]
    \centering
    \caption{Comparison of \gls*{lp}-level and character-level accuracy~(\%) achieved by the GP-\gls*{alpr} and ParSeq-Tiny models on the \dataset dataset for the \gls*{lpr} task. Results represent the average performance over 10 runs using different dataset splits, with standard deviations reported in~parentheses.}
    \label{tab:lpr_results}
    \resizebox{0.8\linewidth}{!}{
        \begin{tabular}{lcc}
            \toprule
            Model & \gls*{lp}-level accuracy &Char-level accuracy \\\midrule
            GP \gls*{alpr} &$93.8~(0.3)$ &$98.7~(0.1)$ \\
            ParSeq-Tiny &$\mathbf{98.0~(0.4)}$ &$\mathbf{99.7~(0.1)}$ \\
            \bottomrule
        \end{tabular}
    }
\end{table}

\begin{figure}[!htb]
    \centering
    \captionsetup[subfigure]{captionskip=1.5pt,labelformat=empty}

    \resizebox{0.99\linewidth}{!}{
        \subfloat[{\ttfamily \stackunder[4pt]{a) AVV3244}{\phantom{a) }\red{M}VV3244}}]{\includegraphics[height=4.5ex]{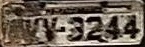}} \hspace{0.15mm}
        \subfloat[{\ttfamily \stackunder[3pt]{b) AEQ1B88}{\phantom{b) }AEQ1B8\red{6}}}]{\includegraphics[height=4.5ex]{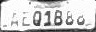}} \hspace{0.15mm}
        \subfloat[{\ttfamily \stackunder[4pt]{c) ANF2072}{\phantom{c) }A\red{M}F2072}}]{\includegraphics[height=4.5ex]{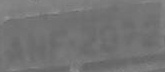}} \hspace{0.15mm}
        \subfloat[{\ttfamily \stackunder[4pt]{d) BDB8C98}{\phantom{d) }BDB\red{2}C98}}]{\includegraphics[height=4.5ex]{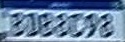}} \hspace{0.15mm}
        \subfloat[{\ttfamily \stackunder[4pt]{e) IMY2G52}{\phantom{e) }I\red{N}Y2G52}}]{\includegraphics[height=4.5ex]{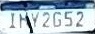}
        }
    }   
    
    \vspace{0.25cm}
    
    \resizebox{0.99\linewidth}{!}{
        \subfloat[{\ttfamily \stackunder[4pt]{f) SFD3H46}{\phantom{f) }S\red{E}D3H46}}]{\includegraphics[height=9.5ex]{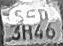}} \hspace{0.15mm}
        \subfloat[{\ttfamily \stackunder[4pt]{g) ASA4083}{\phantom{g) }AS\red{X5}083}}]{\includegraphics[height=9.5ex]{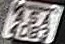}} \hspace{0.15mm}
        \subfloat[{\ttfamily \stackunder[4pt]{h) IKR5H84}{\phantom{h) }\red{AX}R5H\red{6}4}}]{\includegraphics[height=9.5ex]{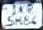}} \hspace{0.15mm}
        \subfloat[{\ttfamily \stackunder[4pt]{i) BBU5364}{\phantom{i) }BBU53\red{5}4}}]{\includegraphics[height=9.5ex]{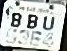}} \hspace{0.15mm}        
        \subfloat[{\ttfamily \stackunder[4pt]{j) SEH2B68}{\phantom{j) }SE\red{R}2B\red{30}}}]{\includegraphics[height=9.5ex]{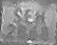}
        }
    } 

    \vspace{1.1mm}
    
    \caption{Example of misrecognized \glspl*{lp}. For each image, the ground-truth label is shown above the model’s prediction, with incorrectly recognized characters highlighted in red. The failure cases include severely degraded characters~(a, h, i), illumination-induced obstructions~(b), low-contrast characters~(c, j), blurring~(d, g), and physically deformed \glspl*{lp}~(e, f).}
    \label{fig:misrecognized_lps}
\end{figure}

Despite the high accuracy, ParSeq-Tiny -- the best-performing method -- still failed in specific scenarios, as shown in \cref{fig:misrecognized_lps}. 
The errors highlight persistent challenges, including highly degraded or low-contrast characters, illumination obstructions, excessive blurring, and physically deformed \glspl*{lp}. 
Low image quality led to confusion between structurally similar characters, such as ``H'' for ``M'' and ``M'' for ``N''. 
Finally, infrared images were a significant source of failure, accounting for \major{$46.2$\%} of all misrecognitions.

\subsection{Joint System Analysis}
\label{subsec:lpr_fgvc_integration}

The previous section showed that baseline \gls*{lpr} methods can produce errors that compromise vehicle identification. 
To address this, this section analyzes the potential for an \gls*{fgvc} system to function as a support mechanism to enhance \gls*{alpr} robustness. 
This analysis establishes a clear path toward unified systems by showing their benefits and challenges.

To evaluate the joint system's performance, we established a clear methodology. First, we used the predictions from the best-performing methods identified in the previous sections: ParSeq-Tiny for \gls*{lpr} and EfficientNet-V2 for \gls*{fgvc} tasks.
Second, we defined a ``correct FGVC set.'' This term is used for metric computation and means that all attributes within a specific combination were predicted correctly for a given image. 
Based on this, we defined three metrics that reflect a practical application scenario.

\begin{itemize}
    \item \NEW{\textbf{Validation Rate} $P(\text{FGVC set correct} \mid \text{LPR correct})$: Measures how often the \gls*{fgvc} support system agrees with a correct \gls*{lpr} prediction.}
    
    \item \NEW{\textbf{Conflict Rate} $P(\text{FGVC set incorrect} \mid \text{LPR correct})$: Quantifies how often the \gls*{fgvc} support system predicts at least one attribute incorrectly, conflicting with the correct \gls*{lpr}.}
    
    \item \NEW{\textbf{Recovery Rate} $P(\text{FGVC set correct} \mid \text{LPR incorrect})$: Measures how often the \gls*{fgvc} support system correctly identifies all vehicle attributes, given the \gls*{lpr} fails.}
\end{itemize}

The results in \cref{tab:integration_analysis} reveal an expected trade-off. 
As more \gls*{fgvc} attributes are required, the Validation and Recovery Rates decrease while the Conflict Rate increases. 
This is a direct consequence of the cumulative error challenge identified in \cref{subsec:fgvc_joint}, amplified by our strict condition that a single attribute error fails the entire \gls*{fgvc} set. 
This methodology explains why the most demanding five-attribute combination yields the worst rates.

\begin{table}[!htp]
    \centering
    \setlength{\tabcolsep}{9pt}
    \caption{Performance metrics of the integrated ALPR-FGVC system. All values are percentages (\%). Validation and Conflict Rates are conditional on a correct LPR; Recovery Rate is conditional on an incorrect LPR. Results are reported as mean (standard deviation in parentheses) over 10 runs.}
    \label{tab:integration_analysis}
    \resizebox{0.99\linewidth}{!}{
        \begin{tabular}{lccc}
            \toprule
            Tasks & \begin{tabular}[c]{@{}c@{}}Validation\\ Rate $\uparrow$\end{tabular} & \begin{tabular}[c]{@{}c@{}}Conflict\\ Rate $\downarrow$\end{tabular} & \begin{tabular}[c]{@{}c@{}}Recovery\\ Rate $\uparrow$\end{tabular} \\
            \midrule
            LPR \& Color                                                            & $93.6$ $(0.5)$ & $\phantom{0}6.5$ $(0.5)$ & $\mathbf{92.7}$ $\mathbf{(0.7)}$\\
            LPR \& Make                                                             & $94.6$ $(0.5)$ & $\phantom{0}5.4$ $(0.5)$ & $85.7$ $(0.8)$\\
            LPR \& Model                                                            & $91.1$ $(0.6)$ & $\phantom{0}8.9$ $(0.6)$ & $82.1$ $(1.4)$\\
            LPR \& Type                                                             & $\mathbf{96.2}$ $\mathbf{(0.6)}$ & $\mathbf{\phantom{0}3.8}$ $\mathbf{(0.6)}$ & $92.1$ $(0.9)$\\
            \midrule
            LPR \& Color \space\& Make                                              & $88.7$ $(0.7)$ & $11.3$ $(0.7)$ & $80.1$ $(0.9)$\\
            LPR \& Color \space\& Model                                             & $85.4$ $(0.8)$ & $14.6$ $(0.8)$ & $76.8$ $(1.4)$\\
            LPR \& Color \space\& Type                                              & $90.2$ $(0.7)$ & $\phantom{0}9.8$ $(0.7)$ & $85.5$ $(0.9)$\\
            LPR \& Make \space\& Model                                              & $88.2$ $(0.8)$ & $11.8$ $(0.8)$ & $75.7$ $(1.3)$\\
            LPR \& Make \space\& Type                                               & $91.8$ $(0.9)$ & $\phantom{0}8.2$ $(0.9)$ & $80.1$ $(1.2)$\\
            LPR \& Model \& Type                                                    & $88.9$ $(1.0)$ & $11.1$ $(1.0)$ & $78.7$ $(1.5)$\\
            \midrule
            LPR \& Color \space\& Make \space\& Model                               & $82.8$ $(0.9)$ & $17.2$ $(0.9)$ & $71.1$ $(1.2)$\\
            LPR \& Color \space\& Make \space\& Type                                & $86.1$ $(1.0)$ & $13.9$ $(1.0)$ & $75.0$ $(1.2)$\\
            LPR \& Color \space\& Model \& Type                               & $83.4$ $(1.0)$ & $16.6$ $(1.0)$ & $73.5$ $(1.4)$\\
            LPR \& Make \space\& Model \& Type                                      & $86.4$ $(1.1)$ & $13.6$ $(1.1)$ & $72.9$ $(1.3)$\\
            \midrule
            LPR \& Color \space\& Make \space\& Model \& Type                       & $81.1$ $(1.1)$ & $18.9$ $(1.1)$ & $68.4$ $(0.7)$\\
            \bottomrule
        \end{tabular}
    }
\end{table}

Despite the trade-off, the results demonstrate the system's powerful capability as a fail-safe. 
The Recovery Rate shows that even in the strictest case, the \gls*{fgvc} system correctly identified a complete vehicle description in $68.4$\% of \gls*{lpr} failures. 
This information is invaluable for practical use, allowing for cross-checking official records or flagging an \gls*{lpr} output as erroneous.

The results also highlight a trade-off between reliability and robustness. 
Relaxing the \gls*{fgvc} set to a single attribute yields high reliability; for instance, the ``LPR + Type'' configuration had the highest Validation Rate ($96.2$\%) and the lowest Conflict Rate ($3.8$\%). 
However, this approach lacks robustness. 
A single attribute like ``Type'' can be misleading, as a misrecognized \gls*{lp} might coincidentally match a database record for a different vehicle sharing the same type. 
In contrast, a multi-attribute set provides a more helpful descriptor (e.g., “Car, Fiat, Uno, White”) that reduces the likelihood of such a false match. 
Therefore, while this multi-attribute check is currently less reliable, it represents the desirable goal for a unified system.

\begin{figure}[!htb]
    \centering
    \captionsetup[subfigure]{captionskip=1.5pt,font=footnotesize}
    \subfloat[Color unk. Fiat Palio]{\includegraphics[height=15ex]{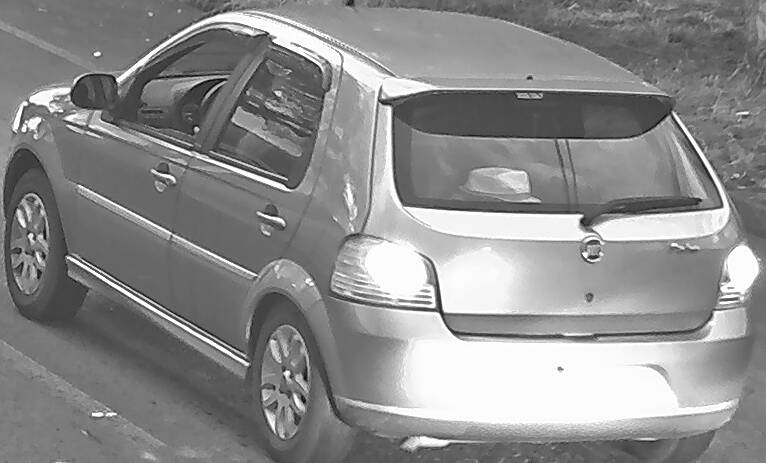}}
    \hspace{0.2mm}
    \subfloat[White Chevrolet Celta]{\includegraphics[height=15ex]{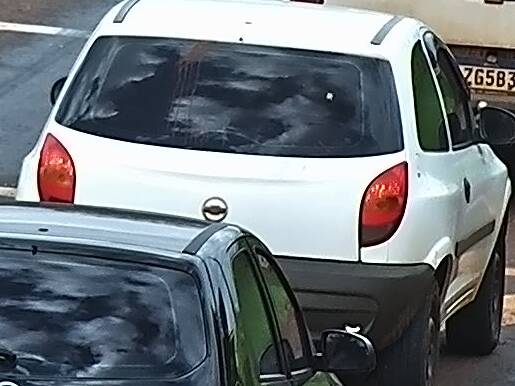}}
    \caption{Examples of \gls{alpr} failures where the \gls{lp} is illegible due to (a)~light glare and (b)~occlusion. Despite these failures, our \gls*{fgvc} classifier acts as a fail-safe, correctly identifying the type, make, and model for both vehicles. Color recognition is correct for~(b), while the infrared image~(a) is classified as ``unknown'' due to the absence of color~information.}
    \label{fig:no_lp}
\end{figure}

Furthermore, \gls*{fgvc} can also function as a fail-safe in scenarios where an \gls*{alpr}-only system would fail completely. 
In real-world surveillance, factors like headlight glare or occlusion can render an LP illegible. 
In these cases, the \gls*{fgvc} system can still be used to analyze the vehicle image and provide valuable information. 
\cref{fig:no_lp} shows illustrative examples of such scenarios.

Finally, we acknowledge this initial analysis has some limitations. 
First, our methodology relied on two simplifications: our ``correct \gls*{fgvc} set'' definition was a strict condition, and we used ground truth to identify errors. 
A practical system should be able to leverage partially correct information and must employ a conflict arbitration method (like confidence scores) to decide whether to trust the \gls*{lpr} or the \gls*{fgvc} outcomes. 
Second, our analysis is dataset-limited and we could not assess the ``false match risk'' -- where an incorrect \gls*{lpr} might coincidentally retrieve a vehicle record that matches the \gls*{fgvc} output. 
Evaluating this risk requires access to an entire official vehicle database (including \gls*{lp} and attribute records) and remains a key direction for future~research.

%% file: 7-conclusion.tex
\section{Conclusions}
\label{sec:Conclusion}

In this work, we introduce \dataset, a public dataset for \gls*{fgvc} and \gls*{alpr} research in surveillance scenarios. 
It contains annotations for $13$ vehicle colors, $26$ manufacturers, $136$ models, and $14$ types (all validated against official records), plus \gls*{lp} text labels and corner annotations. 
A quantitative and qualitative analysis confirms the dataset captures challenging real-world conditions. 
As the first of its kind, \dataset supports both independent and combined research in \gls*{fgvc} and \gls*{alpr}.

Benchmark experiments for vehicle color, make, model, and type recognition were conducted, with EfficientNet-V2 achieving the best overall performance. 
All recognition tasks achieved micro-accuracy scores above $90$\%. 
Despite this performance, an error analysis revealed remaining challenges that warrant further attention, including infrared images, multicolored vehicles, and distinguishing similar body-style model variants.

Furthermore, our analysis of joint \gls*{fgvc} tasks revealed that combining attributes reduces simultaneous recognition rates and leads to inconsistent predictions, such as a correct model with an incorrect make. 
A promising solution is to replace isolated classifiers with a multi-task learning~\citep{caruana1997multitask} framework. 
This approach can enforce hierarchical consistency and leverage natural correlations to enhance generalization across tasks.

We also benchmarked two methods for the \gls*{lpr} task. 
ParSeq-Tiny achieved the highest recognition rate, exceeding $98$\%. 
Nonetheless, error analysis revealed areas for improvement, particularly in handling distorted or physically deformed \glspl*{lp}.
Building on this, we explored the combined use of \gls*{fgvc} and \gls*{lpr}, demonstrating that \gls*{fgvc} functions as fail-safe. 
Our results showed that even in the strictest case, the \gls*{fgvc} system provided a complete, correct vehicle description for $\approx68$\% of \gls*{lpr} failures.

Note that the \dataset dataset's focus on high quality \gls*{fgvc} annotations -- relying on official records retrieved from \gls*{lp} information -- means it does not fully capture the challenges of unconstrained \gls*{alpr}. 
In such scenarios, systems face significant legibility issues, as over $25$\% of images are inadequate for recognition~\citep{wojcik2025lplc}. 
This further highlights the importance of our proposed \gls*{alpr}-\gls*{fgvc} integration, which can help validate correct recognitions or reject unreliable outputs. 
Both actions are crucial for reducing false positives.

Future research is needed to develop methods for weighing predictions from different recognition systems and assessing their reliability. 
A promising solution is selective prediction~\citep{geifman2017selective}, but this necessitates classifier calibration, as standard confidence scores are often overconfident~\citep{guo2017calibration,minderer2021revisiting,laroca2023leveraging}. 
Extending these ideas to top-k settings also raises open questions, such as how to handle illogical Make–Model combinations in joint \gls*{fgvc} tasks or how to define top-k predictions in \gls*{alpr} (e.g., at the character level or for entire sequences). 
A thorough exploration of these issues remains an important direction for future~work.

Looking ahead, we plan to release a large-scale dataset inspired by \dataset, comprising over a million surveillance images with greater diversity and more unconstrained conditions -- especially for advancing \gls*{alpr} research in such scenarios. 
This new dataset will likely feature semi-automatically generated annotations, highlighting the need to explore strategies for effectively training models with noisy or coarse labels~\citep{lucio2019simultaneous}, while striking a balance between annotation efficiency and recognition performance.

%% file: 0-acknowledgments.tex
\section*{Acknowledgments}

This study was partially funded by the \textit{Coordenação de Aperfeiçoamento de Pessoal de Nível Superior - Brasil~(CAPES)} -- Finance Code 001, and by the \textit{Conselho Nacional de Desenvolvimento Científico e Tecnológico~(CNPq)} (\#~315409/2023-1 and \#~312565/2023-2).
We gratefully acknowledge the support of NVIDIA Corporation with the donation of the Quadro RTX $8000$ GPU, which was used in part of this research.
We also extend our sincere thanks to students Leonardo Carpwiski, Rafael Marques, and Thiago Assunção for their assistance in labeling the \gls*{lp} texts in the \dataset~dataset.